\documentclass[journal,twoside,web]{ieeecolor}
\usepackage{tmi}
\usepackage{cite}
\usepackage{amsmath,amssymb,amsfonts}
\usepackage{algorithmic}
\usepackage{booktabs}
\usepackage[ruled,noend,linesnumbered]{algorithm2e}
\usepackage{multirow,makecell}
\newcommand{\tabincell}[2]{\begin{tabular}{@{}#1@{}}#2\end{tabular}}

\usepackage{xcolor}

\usepackage{bbding}
\usepackage{tikz}
\usepackage{graphicx}
\usepackage{subcaption}
\usepackage{textcomp}
\def\BibTeX{{\rm B\kern-.05em{\sc i\kern-.025em b}\kern-.08em
    T\kern-.1667em\lower.7ex\hbox{E}\kern-.125emX}}
\begin{document}
\title{Cell Instance Segmentation: \\
The Devil Is in the Boundaries}
\author{Peixian Liang, Yifan Ding, Yizhe Zhang, Jianxu Chen, Hao Zheng,
Hongxiao Wang, Yejia Zhang, \\ Guangyu Meng, Tim Weninger, Michael Niemier, X. Sharon Hu, Danny Z Chen 
\thanks{This research was supported in part by NSF grant 2212239, DARPA contracts HR001121C0168 and HR00112290106, and the Louisiana Board of Regents under Contract Number LEQSF(2025-28)-RD-A-21. The work of J.C. was partially supported by the “Ministerium für Kultur und Wissenschaft des Landes Nordrhein-Westfalen” and “Der Regierende Bürgermeister von Berlin, Senatskanzlei Wissenschaft und Forschung”, and by the Bundesministerium für Forschung, Technologie und Raumfahrt, BMFTR under the funding reference 161L0272.}
\thanks{Peixian Liang, Yifan Ding, Yizhe Zhang, Hongxiao Wang, Yejia Zhang, Guangyu Meng, Tim Weninger, Michael Niemier, X. Sharon Hu, and Danny Z. Chen are with the Department of Computer Science and Engineering, University of Notre Dame, Notre Dame, IN 46556, USA (e-mail: \{pliang, yding4, yzhang29, hwang21, yzhang46, gmeng, tweninge, mniemier, shu, dchen\}@nd.edu). }
\thanks{Jianxu Chen is with Leibniz-Institut für Analytische Wissenschaften–ISAS–e.V., Dortmund 44139, Germany (e-mail: jianxu.chen@isas.de).}
\thanks{Hao Zheng is with the School of Computing and Informatics, University of Louisiana at Lafayette, Lafayette, LA 70503, USA (email: hao.zheng@louisiana.edu).}
}

\maketitle

\begin{abstract}
State-of-the-art (SOTA) methods for cell instance segmentation are based on deep learning (DL) semantic segmentation approaches, focusing on distinguishing foreground pixels from background pixels. In order to identify cell instances from foreground pixels (\textit{e.g.}, pixel clustering), most methods decompose instance  
information into pixel-wise objectives, such as 
distances to foreground-background boundaries (distance maps), 
heat gradients with the center point as heat source (heat diffusion maps), and distances from the center point to foreground-background boundaries with fixed angles (star-shaped polygons). 
However, pixel-wise objectives may lose significant geometric properties of the cell instances, such as shape, curvature, and convexity, which require a 
collection of pixels to represent. To address this challenge, we present a novel pixel clustering method, called {\textbf Ceb} (for {\textbf Ce}ll {\textbf b}oundaries), to leverage cell boundary features and labels to divide foreground pixels into cell instances. Starting with probability maps generated from 
semantic segmentation, Ceb first extracts potential foreground-foreground boundaries (\textit{i.e.}, boundary candidates) with a revised Watershed algorithm. For each boundary candidate, a boundary feature representation (called boundary signature) is constructed by sampling pixels from the current foreground-foreground boundary as well as the neighboring background-foreground boundaries. Next, a lightweight boundary classifier 
is used to predict its binary boundary label based on the corresponding boundary signature. Finally, cell instances are obtained by dividing or merging neighboring regions based on the predicted 
boundary labels. Extensive experiments on
six 
datasets demonstrate that Ceb outperforms existing 
pixel clustering methods on semantic segmentation probability maps. Moreover, Ceb achieves highly competitive performance compared to state-of-the-art cell instance segmentation methods. The code is available at: https://github.com/pxliang/Ceb.
\end{abstract}


\section{Introduction}
\label{sec:introduction}


\IEEEPARstart{C}{ell} instance segmentation is a fundamental problem in quantitative cell biology research. It provides accurate and detailed information about individual cells, including their positions, morphology, and life cycle. This level of granularity is crucial for understanding cellular behaviors, dynamics, and interactions~\cite{madukoma2019single}. 
Unlike other instance segmentation tasks such as those in natural scenes (e.g., the COCO dataset~\cite{lin2014microsoft}) and multi-class organs (e.g., heart~\cite{zhuang2016multi} and chest~\cite{shiraishi2000development}), cell instance segmentation presents distinguished characteristics and challenges. For example, cell instances typically exhibit a roughly convex shape (e.g., resembling a star-shaped polygon whose entire boundary is visible from an interior point). A single image of cells can contain hundreds or even thousands of individual cells. In many scenarios, cells are tightly packed together. Cells of the same type can have either similar or quite different sizes and different textual characteristics. 

\begin{figure*}[ht]
\centering
\includegraphics[width=0.8\textwidth]{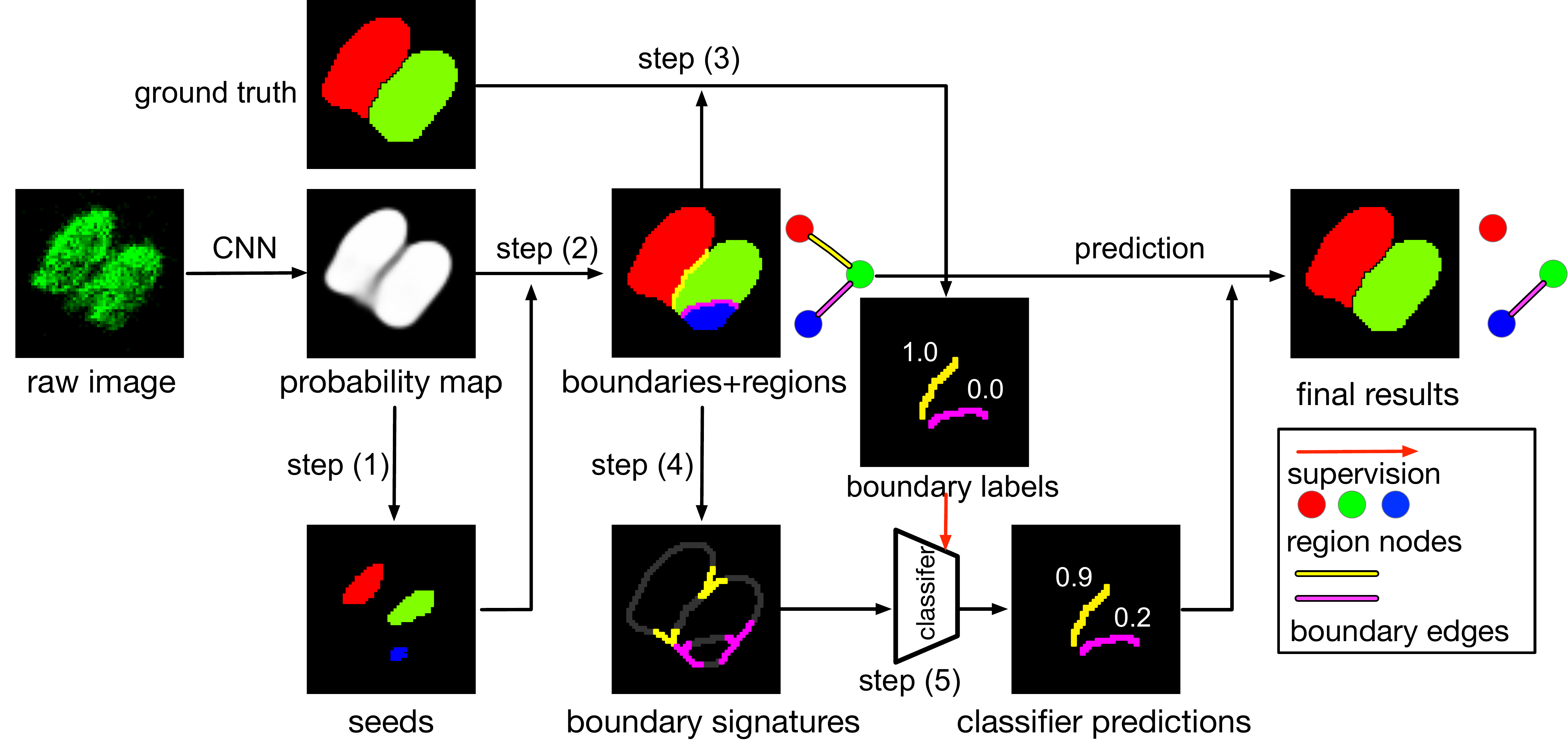}
\caption{An overview of our Ceb framework. An input image is fed to a CNN network (\textit{e.g.}, U-Net) to produce a semantic probability map. \textbf{Step (1) seed generation} generates seeds from the probability map. \textbf{Step (2) boundary generation} uses these seeds and the probability map to produce possible boundaries and the divided regions. \textbf{Step (3) boundary label assignment} matches ground truth instance masks and the divided regions to attain true regions and their corresponding boundaries (as true boundaries).
\textbf{Step (4) boundary signature extraction} generates boundary-based feature representations, boundary signatures, for all possible boundaries. During training, these boundary signatures, along with their corresponding true/false boundary labels obtained in \textbf{Step (3)}, are fed to \textbf{Step (5) boundary classification} to train a boundary classifier. 
During inference, the boundary classifier predicts a true/false label for each possible boundary based on its boundary signature. The final instance results are obtained by merging connected regions divided by false boundaries.}
\label{fig:overview}
\end{figure*}

Deep learning (DL) semantic segmentation methods have exhibited remarkable performance in biomedical image segmentation tasks. 
Existing state-of-the-art (SOTA) methods for cell instance segmentation rely on semantic segmentation to distinguish foreground pixels from background pixels. In order to identify accurate cell instances among foreground pixels (\textit{e.g.}, pixel clustering), most current methods use pixel-wise objectives. 
Hover-Net~\cite{graham2019hover} used shortest distances to foreground-background boundaries (distance maps); CellPose~\cite{stringer2021cellpose} used heat gradients with the center point as heat source (heat diffusion maps); StarDist~\cite{schmidt2018cell} used multiple distances from inner point to foreground-background boundaries with fixed angles (star-shaped polygons); pixel-embedding methods~\cite{payer2019segmenting,zhao2021faster,lalit2022embedseg,goldsborough2024instanseg} used learnable vectors to indicate pair-wise similarities (pixel embeddings). However, these pixel-wise objectives ignore significant
geometric properties of the original cell instances, such as
shape, curvature, and convexity, which require a structured collection of
pixels to represent. 

In this paper, we propose a new approach for clustering foreground pixels into cell instances, called Ceb (for {\textbf Ce}ll {\textbf b}oundaries), based on boundary-wise features and labels. Unlike pixel-wise objectives, boundary-wise features and objectives are based on a structured group of key pixels, and thus instance geometric properties can be better preserved. 
Specifically, our Ceb framework work as follows.
Given a probability map produced by a DL semantic segmentation model ({\textit{e.g.}}, U-Net \cite{ronneberger2015u}) for an input image, we first compute 
instance seeds from the probability map, and  
generate all potential cell boundaries with a revised Watershed algorithm~\cite{meyer1994topographic}. 
In the training stage, labels (true or false) of the binary boundaries thus generated are obtained by computing optimal matching between ground truth cell instances and all the possible instance candidates produced by enumerating all possible binary boundary labels. For each boundary candidate, we then extract a novel boundary feature representation (called boundary signature) by sampling key pixels from the current foreground-foreground boundary as well as the neighboring background-foreground boundaries. Based on the extracted boundary signatures, we use a lightweight binary image classifier (a convolutional neural network (CNN)) to distinguish true and false boundaries. In the inference stage, positive boundaries are kept, negative boundaries are removed, and all neighboring regions divided by predicted negative boundaries are merged to form cell instances. On 2D temporal (video) datasets, our method can further incorporate temporal consistency information~\cite{liang2022h} to better segment cell instances.

We conduct experiments on six cell instance segmentation datasets (four video datasets and two 2D datasets). Our boundary classifier shows an excellent ability to distinguish true/false boundaries based on novel boundary-based features. Ceb outperforms all the compared foreground pixel clustering methods on semantic probability maps across different datasets. Compared to SOTA instance segmentation methods, Ceb also yields competitive performances. In video settings, our temporal-consistency based method further improves the performance.
In summary, our main contributions are as follows.

\begin{itemize}
\item  We propose Ceb, a novel method for clustering foreground pixels into cell instances. Ceb tackles the pixel clustering problem with a DL boundary classifier based on boundary-level features, which can better preserve instance geometric properties compared to existing pixel-wise objectives.

\item  We develop a novel boundary-level feature representation, called boundary signature, by sampling pixels from each potential foreground-foreground boundary and its neighboring background-foreground boundaries. Boundary signatures can effectively reflect geometric properties which are essential for distinguishing true and false boundaries.

\item  We revise the Watershed algorithm to generate all potential foreground-foreground boundaries, and present an optimized instance matching method to assign labels to the generated boundaries. 

\item We propose a novel matching method to incorporate temporal instance consistency into the Ceb framework, further improving instance segmentation performance on 2D temporal datasets.

\end{itemize}

\section{Related Work}

\subsection{Cell Instance Segmentation}

Recent SOTA methods for cell instance segmentation can be broadly categorized into two types: semantic segmentation-based approaches and region-based approaches, with the vast majority of SOTA methods belonging to the former type.

\subsubsection{Semantic Segmentation-based Approaches}

Semantic segmentation-based methods first distinguish foreground pixels and background pixels, and then cluster foreground pixels into individual instances~\cite{chen2016dcan,graham2019mild,zhou2019cia,ronneberger2015u,isensee2021nnu,liang2019cascade}. The most recent work adopted pixel-wise objectives to distinguish individual instances. In the training stage, cell instance labels are transformed to pixel-wise labels. In the inference stage, pixel-wise labels are predicted and further processed to produce final instances. For example, Hover~\cite{graham2019hover} and CellViT~\cite{horst2024cellvit} used distance maps considering the distances from each inner pixel to the nearest instance boundaries.  
StarDist~\cite{schmidt2018cell} extended distance maps to radial maps with $32$ fixed angles. CellPose~\cite{stringer2021cellpose} introduced heat diffusion maps with the center point as the heat source, and the heat gradients of each point were considered as pixel-wise labels. Another kind of popular methods is based on pixel-embedding, using contrastive learning to represent pixel-pixel similarities~\cite{payer2019segmenting,zhao2021faster,lalit2022embedseg,goldsborough2024instanseg, wang2024mudslide}. Pixels with similar embeddings are clustered together to form the final instance segmentation results. For example, InstanSeg~\cite{goldsborough2024novel} predicted seed points that represent instance centers and learned pixel-wise embeddings which were used to cluster pixels into instances based on their similarity to seed embeddings. Despite their flexibility and generalizability, these pixel-wise objectives may still lose significant geometric properties of original cell instances, such as
shape, curvature, and convexity, which require a structured collection of pixels to represent. In this work, we present Ceb, aiming to preserve the structure properties of cell instances with boundary-level features by selecting pixels from foreground-foreground boundaries and background-foreground boundaries.

\subsubsection{Region-based Approaches}

In region-based models, instances are assigned to grids or anchors within an image, allowing for region-wise classification to detect and segment instances~\cite{he2017mask,liu2018path,chen2019hybrid,huang2019mask,cai2019cascade,tang2021look}. For example, CelloType~\cite{pang2024cellotype} employed a Transformer-based detector (DINO) to generate bounding boxes and extract latent features, which are then processed by MaskDINO for joint optimization of detection, segmentation, and classification. However, such region-based representations may not be a good fit for cell instances. For example, bounding boxes can be suppressed by nearby instances, especially in crowded scenes. Pre-defined bounding boxes can also suffer from the imbalance problem of false/true instances.



\subsection{Boundary Generation Methods}
Instance boundaries play a vital role in image segmentation.
Traditional boundary generation methods generate instance boundaries based on local pixel features. Two well-known methods are Watershed~\cite{meyer1994topographic} and Active Contours~\cite{caselles1997geodesic}. The Watershed algorithm considers an image as a heat map based on pixel intensities. Instance boundaries are determined as local minimum lines/curves between instances. Subsequent work incorporated the Watershed algorithm with DL methods~\cite{lux2019dic,eschweiler2019cnn}. For example, DIC~\cite{lux2019dic} predicted cell seeds first and used Watershed as a foreground pixel clustering step to obtain final cell instances. In~\cite{wolf2017learned}, the Watershed algorithm is transformed into a learnable model to consider altitudes of pixels as well as the corresponding region assignment. Active contours are energy-minimizing curves that deform and converge to the boundaries of the regions of interest (RoIs) in an image~\cite{kass1988snakes,chan2001active,sundaramoorthi2007sobolev}. In~\cite{rupprecht2016deep}, parameter maps/initial contours for Sobolev active contours were predicted by CNN models, and Sobolev active contours were applied as a foreground pixel clustering step to obtain final predictions. Subsequent work proposed active contour inspired losses (e.g., Mumford-Shah loss~\cite{kim2019mumford}, active contour without edge (ACWE) loss~\cite{chen2019learning,zhang2020deep, kim2019cnn, gur2019unsupervised}, and sneak active contour loss~\cite{marcos2018learning}). Boundary-based approaches are also a rising focus for point cloud segmentation. For example, CBL~\cite{tang2022contrastive} proposed a boundary contrastive loss for point cloud segmentation.
Unlike most existing boundary-based methods which utilize solely pixel-level features to generate boundaries, Ceb employs boundary-level features to classify binary boundary labels on top of foreground pixels. Consequently, Ceb retains the advantages of semantic segmentation compared to the other boundary generation methods while still being able to preserve instance structure properties as the known boundary generation methods.


\subsection{Segmentation Trees}
Another line of related work is segmentation tree based methods~\cite{silberman2014instance,funke2018candidate,fehri2019bayesian,souza2016overview,jones1999connected,akram2014segmentation, akram2016joint}, which utilize tree-based structures to solve instance segmentation problems. These methods generate over-segmented components, and then perform final instance segmentation by clustering the components. Unlike our method, such methods often produce tree-like structures without DL networks. For example, a tree can be generated from super-pixels~\cite{silberman2014instance, funke2018candidate}; after the ``leaf" regions of the initial over-segmented candidate regions are obtained, a tree structure is built by iteratively merging similar super-pixels, until a pre-specified stopping criterion is met. GP-S3Net~\cite{razani2021gp} used density-based spatial clustering of applications with noise (DBSCAN)~\cite{ester1996density} to produce over-segmented candidates, and a graph neural network (GNN) model was used to predict the label of each edge. Note that these methods make final decisions based on node features, which make modeling geometric features of instance regions difficult. In contrast, our method directly utilizes boundary features for boundary classification, which are generally easier to identify and classify.

\begin{algorithm}
\footnotesize
\SetAlgoLined
\SetKw{KwInput}{Input}
\SetKw{KwOutput}{Output}
\KwInput{Foreground Pixels $F = \{ (x_{f}, y_{f})\}$;
Seeds $S = \{ (x_s, y_s)\} \in F$;
Probability map $p$, $p(x_f, y_f) \in [0, 1], \forall (x_f, y_f) \in F$;
}  

\KwOutput{Regions $\boldsymbol{R}$, Boundaries $\boldsymbol{B}$ 
\# \textit{Regions and Boundaries are both Hashmaps, representing index and corresponding pixels. }
}

$\boldsymbol{R} \gets \emptyset, \boldsymbol{B} \gets \emptyset$ 

\# \textit{status mapping function $f$: $-1$ represents unvisited ($\textbf{UNVISITED}$), $-2$ represents in the queue to be assigned ($\textbf{INQE}$), $-3$ represents a place holder to be placed in the queue later ($\textbf{MASK}$), $0$ represents a Watershed boundary ($\textbf{WSHD}$), positive integer represents the region index.}

\For{$(x, y) \in F$}{
    $f[(x, y)] \gets -1$
}

$region\_index \gets 0$

\For{$(x, y) \in S$}{
    $region\_index \gets region\_index + 1$
    
    $f[(x, y)] \gets region\_index$
    
    $\textbf{R}[f[(x, y)]].add((x, y))$
}

Initialize $Queue \gets \emptyset$

Initialize $Hashmap \gets \emptyset$ \# \textit{key represents probability map value, value is a list of pixels with the corresponding probability}

\For{$(x, y) \in$ $F \setminus S$ }{
    $Hashmap[p(x, y)].add((x, y)) $
}

 \For{$key \in reverse\_sorted(Hashmap.keys)$}{
   \For{$(x, y) \in Hashmap[key]$}{
        $f[(x, y)] \gets \textbf{\textit{MASK}}$
        
        \For{$(x_n, y_n)\in Neighbors(x, y)$}{
            \If{$f[(x_n, y_n)] > 0$ }{ 
                $Enqueue(Queue, (x, y))$
                $f[(x, y)] \gets$ \textbf{\textit{INQE}}
                
                \textbf{break}
                }
  }
  }
 \While{$Queue \neq \emptyset$}{
   $(x, y) \gets Dequeue(Queue)$\;
   \For{$(x_n, y_n) \in Neighbors(x, y)$}{
            \If{$f[(x_n, y_n)] > 0$}{
               \If{$f[(x, y)] = \textbf{INQE}$}{
                    $f[(x, y)] \gets f[(x_n, y_n)]$
                    $\textbf{R}[f[(x_n, y_n)]].add((x, y))$
                    }
                \uElseIf{$f[(x, y)]>0$ \textbf{and} $f[(x, y)] \neq f[(x_n, y_n)]$}{
                $\textbf{B}[f[(x, y)], f[(x_n, y_n)]].add((x, y))$
                $f[(x, y)]\gets \textbf{\textit{WSHD}}$
                }
                }
            \uElseIf{$f[(x_n, y_n)] = \textbf{\textit{WSHD}}$}{
             \If{$f[(x, y)] = \textbf{INQE}$}{
                    $\textbf{B}[ \textbf{B}.find\_key((x_n, y_n)) ].add((x, y))$
                    $f[(x, y)] \gets \textbf{\textit{WSHD}}$
                    }
            }
            \uElseIf{$f[(x_n, y_n)] = \textbf{MASK}$}{
                $f[(x_n, y_n)] \gets \textbf{\textit{INQE}}$
                $Enqueue(Queue, (x_n, y_n))$
            }
                }
 }}
 \Return{$\boldsymbol{B}$, $\boldsymbol{R}$}
\caption{A Revised Watershed Algorithm to Generate Possible Cell Boundaries and Regions}
\label{algorithm:adapted_watershed}
\end{algorithm}


\section{Methodology}
\label{method}

In this section, we present our Ceb framework for cell instance segmentation. Fig.~\ref{fig:overview} gives an overview of our framework, which consists of five main steps. (1) Seed generation (Section~\ref{section:seed_generation}): We first generate instance seeds from semantic segmentation probability maps. (2) Boundary generation (Section~\ref{section:boundary_generation}): Given the generated seeds and probability maps, we employ a revised Watershed algorithm to generate possible cell boundaries and the regions enclosed by these boundaries. Thus, the cell instance segmentation problem becomes a boundary selection problem. (3) Boundary label assignment (Section~\ref{section:boundary_label_assignment}): To obtain boundary labels (true or false) for the training stage, we build an optimized matching model 
between ground truth instance masks and the divided regions to attain true regions. The corresponding boundaries of true regions are true boundaries while the other boundaries are false boundaries. (4) Boundary signature extraction (Section~\ref{section:boundary_signature_extraction}): A novel type of boundary features, called boundary signature, is extracted for each boundary to capture its geometric characteristics. Boundary signatures serve as input for the subsequent step of boundary classification. (5) Boundary classification (Section~\ref{section:boundary_classification}): We build a lightweight binary boundary classifier based on the extracted boundary signatures and the corresponding boundary labels. The final cell instances are obtained by keeping true boundaries and merging connected regions separated by false boundaries. 

For cell video datasets that have extra properties of temporal instance consistency, we further incorporate such instance consistency information into our method. Specifically, we incorporate both temporal instance consistency and boundary probability scores by the boundary classifier to produce final instance segmentation using a matching and selection method.

\subsection{Seed Generation}
\label{section:seed_generation}
In this step, we 
generate seeds using the probability map, which are then fed to the next boundary generation step (see step (1) in Fig.~\ref{fig:overview}).
The seeds are generated from an instance candidate forest (ICF)~\cite{liang2022h}. The process is as follows: Given an input image $x$ with a foreground probability map $p$ from a pixel-wise classification model, the probability value of each pixel ranges from 0 to 1. All the probability values of $p$ are sorted into a list $V = \{v_1, v_2, \ldots,v_H \}$ in increasing order after removing duplicated values and merging highly similar values. Each value $v_h \in V$ as a threshold value determines the connected components in $x$ whose pixels' probability values are all $\geq v_h$. Thus, $v_h$ induces an instance candidate set $C^{v_h}$, which is a collection of mutually disjoint regions in $x$. The pixels of instance candidates in the set $C^{v_h}$ are a subset of the pixels of instance candidates in the set $C^{v_{h-1}}$ for $h>1$. Thus, we have a forest structure $\mathcal{F} = (C, \mathcal{H})$, where $C$ is the node set of all the candidate regions and $\mathcal{H}$ is the parent function for each node in $C$. We then select the local maximal values of all the leaf nodes in $\mathcal{F}$ as the set $S$ of seeds. 

\subsection{Boundary Generation}
\label{section:boundary_generation}
Given the probability map $p$ and the set $S$ of seeds, we seek to generate all possible cell boundaries and the corresponding regions for the connected components of foreground pixels (see step (2) in Fig.~\ref{fig:overview}). One can expect that each 
boundary is adjacent to two different regions.
Let $B$ be the set of boundaries  
and $R$ be the set of regions thus obtained.

We revised the Watershed algorithm~\cite{meyer1994topographic} to generate possible boundaries (called region-region boundaries in Section~\ref{section:boundary_signature_extraction}). In a high-level view, Watershed is a seed-growth method based on pixel intensities. First, each seed is initialized as an individual instance. Then, all the other pixels are ordered by their intensities and assigned labels by the neighboring pixels. If a pixel is attached only to pixels of a single instance, it is labeled as the same instance. If a pixel is attached to multiple instances, it is labeled as a boundary (Watershed line) and indexed by the labels of the attached instances. If a pixel is attached only to one pixel of a boundary, it is labeled as the same boundary. See Algorithm~\ref{algorithm:adapted_watershed} for more details.  

\begin{figure}[t]
\centering
\includegraphics[width=0.5\textwidth]{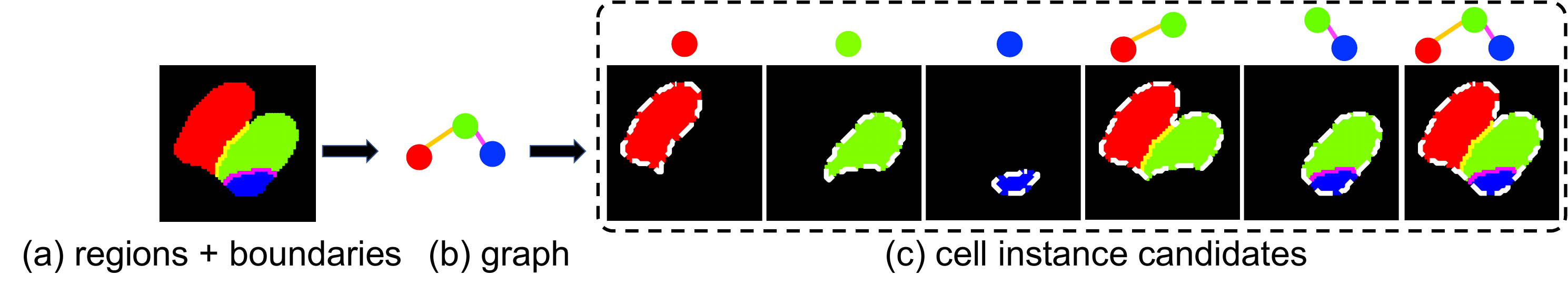}
\caption{The process of generating cell instance candidates from the possible regions and boundaries. Given the boundaries and regions (a), an undirected graph is built (b), in which the regions are represented as nodes and boundaries as edges. By enumerating all possible connected subgraphs, all instance candidates are obtained (c).
}
\label{fig:region_to_ins}
\end{figure}

As illustrated in step (2) of Fig.~\ref{fig:overview}, the regions divided by the possible boundaries can be modeled as an undirected graph, $G = (R, B)$, by considering each individual region as a node and each generated boundary as an edge between two attached regions. If a subgraph $G_s \subseteq G$ is a connected graph, its corresponding node set $I_s$ can form a possible cell instance. All the possible cell instances in $G$ constitute an instance candidate set $\mathcal{I} = \{ I_s \}$ (see Fig.~\ref{fig:region_to_ins}).

\subsection{Boundary Label Assignment}
\label{section:boundary_label_assignment}
Given the generated boundaries, we aim to assign them true or false labels for the training stage. Note that a boundary is generated from the probability map $p$, and thus is quite likely different from the boundaries of the ground truth masks. To assign true/false boundary labels, we formulate it as a problem of computing optimal instance matching between ground truth instances and all the possible valid instances of the prediction. 

\noindent
\textbf{Boundary Label Assignment Matching Model.}
This matching model aims to find an optimal matching flow between ground truth instances $\mathcal{G}$ and the instance candidate set $\mathcal{I}$. 
\begin{equation}\label{equation:object_function_1}
  \textbf{GI-matching}(\mathcal{G}, \mathcal{I}) = \max_{f} \sum_{i \in \mathcal{G}} \sum_{j \in \mathcal{I}} M_{i,j} f_{i,j}
\end{equation}
\begin{equation}\label{equation:constraint_1-1}
    \sum_{j \in \mathcal{I}} f_{i,j} \leq 1, \forall i \in \mathcal{G},
\end{equation}
\begin{equation}\label{equation:constraint_1-2}
\sum_{i\in \mathcal{G}}\sum_{k \in K(r)} f_{i,k} \leq 1, \forall r \in R,
\end{equation}
\begin{equation} \label{equation:constraint_1-3}
     f_{i,j} \in \{0,1\}, \forall i \in \mathcal{G}, \forall j \in \mathcal{I}.
\end{equation}

\begin{figure}[t]
\centering
\includegraphics[width=0.5\textwidth]{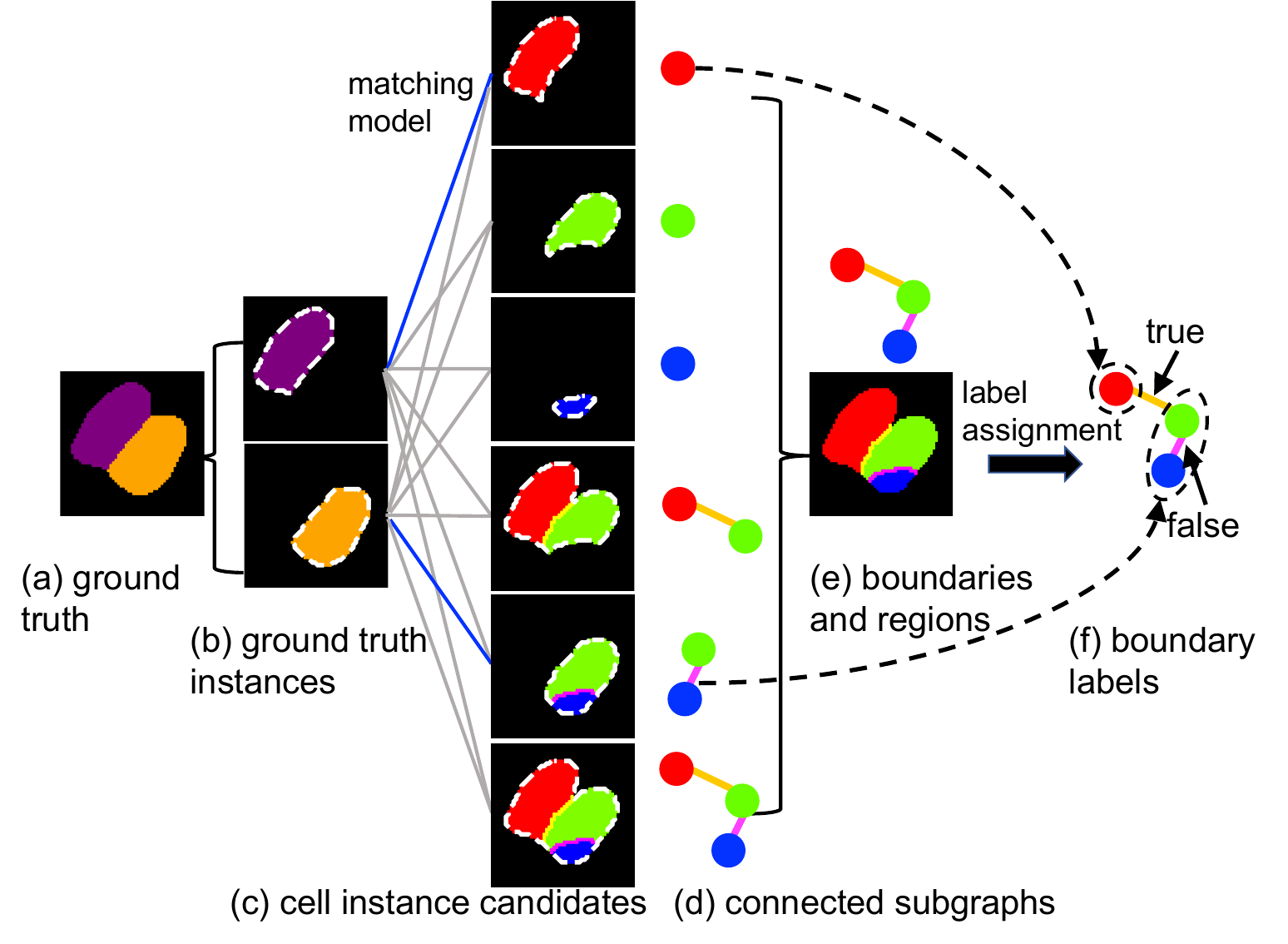}
\caption{The optimal instance matching model for boundary label assignment.
We consider instance-level matching between ground truth (GT) instance masks (a) and the generated boundaries and regions (e). The GT labels are decomposed into individual ground truth instances (b); possible cell instance candidates (c) are generated by considering all possible connected subgraphs (d) in the graph (e). Dashed lines indicate the matching results (two instances are selected). The boundaries inside the matched instances are assigned false labels, and those enclosing the matched instances are assigned true labels. 
}
\label{fig:matching_model}
\end{figure}

\noindent
The objective of the above matching model is to maximize the sum of all the matching scores $M_{i, j} \in M$ multiplied by the flow variables $f_{i,j} \in \{0, 1\}$, where $i \in \mathcal{G}$ represents a ground truth (GT) instance mask and $j \in \mathcal{I}$ is a possible cell instance. We use the measure of Intersection-over-Union (IoU) for matching scores in $M$. We require each GT mask to match with at most one instance candidate (see Eq.~(\ref{equation:constraint_1-1})). Further, each region $r \in R$ can be selected at most once. This constraint is enforced for each region $r\in R$ by considering all possible instance candidates that contain $r$, that is, $K(r)$ is the collection of all possible instance candidates that contain $r$. For each $r\in R$, we require that the sum of all the flows to the regions in $K(r)$ be less than or equal to $1$ (see Eq.~(\ref{equation:constraint_1-2})). This optimal matching model is solved by integer linear programming (ILP). Then for each instance candidate $j \in \mathcal{I}$ $(\sum_{i \in \mathcal{G}}f_{i, j} = 1)$ which is selected by the model, all its internal boundaries are taken as false boundaries, and all the other boundaries are taken as true boundaries (see Fig.~\ref{fig:matching_model}).

\begin{algorithm}
\SetAlgoLined
\footnotesize
\SetKw{KwInput}{Input:}
\SetKw{KwOutput}{Output:}
\KwInput{Regions $\boldsymbol{R}$, Boundaries $\boldsymbol{B}$.
}  

\KwOutput{Boundary Signature $\boldsymbol{BS}(b)$}.

Obtain foreground-background boundaries {$\boldsymbol{\hat{{B}}}$};
$\boldsymbol{\hat{B}} = \boldsymbol{B} \cup \boldsymbol{\hat{{B}}}$;

\For{$b \in \boldsymbol{B}$}{
    
    $(n_1, n_2) \gets find\_endpoints(b)$;

    $(\hat{b}_{11}, \hat{b}_{12}) \gets nearest\_boundaries(n_1, \boldsymbol{\hat{B}} \setminus b)$;

    $(\hat{b}_{21}, \hat{b}_{22}) \gets nearest\_boundaries(n_2, \boldsymbol{\hat{B}} \setminus b)$;

    $\boldsymbol{BS}(b) = sample\_points(b,  \hat{b}_{11},  \hat{b}_{12},  \hat{b}_{21},  \hat{b}_{22}, n_1, n_2)$;
    
}

\Return{$\boldsymbol{BS}(b)$}.
\caption{Boundary Signature Extraction}
\label{algorithm:boundary_signature_extraction}
\end{algorithm}

\subsection{Boundary Signature Extraction}
\label{section:boundary_signature_extraction}
\begin{figure}[t]
\centering
\includegraphics[width=0.45\textwidth]{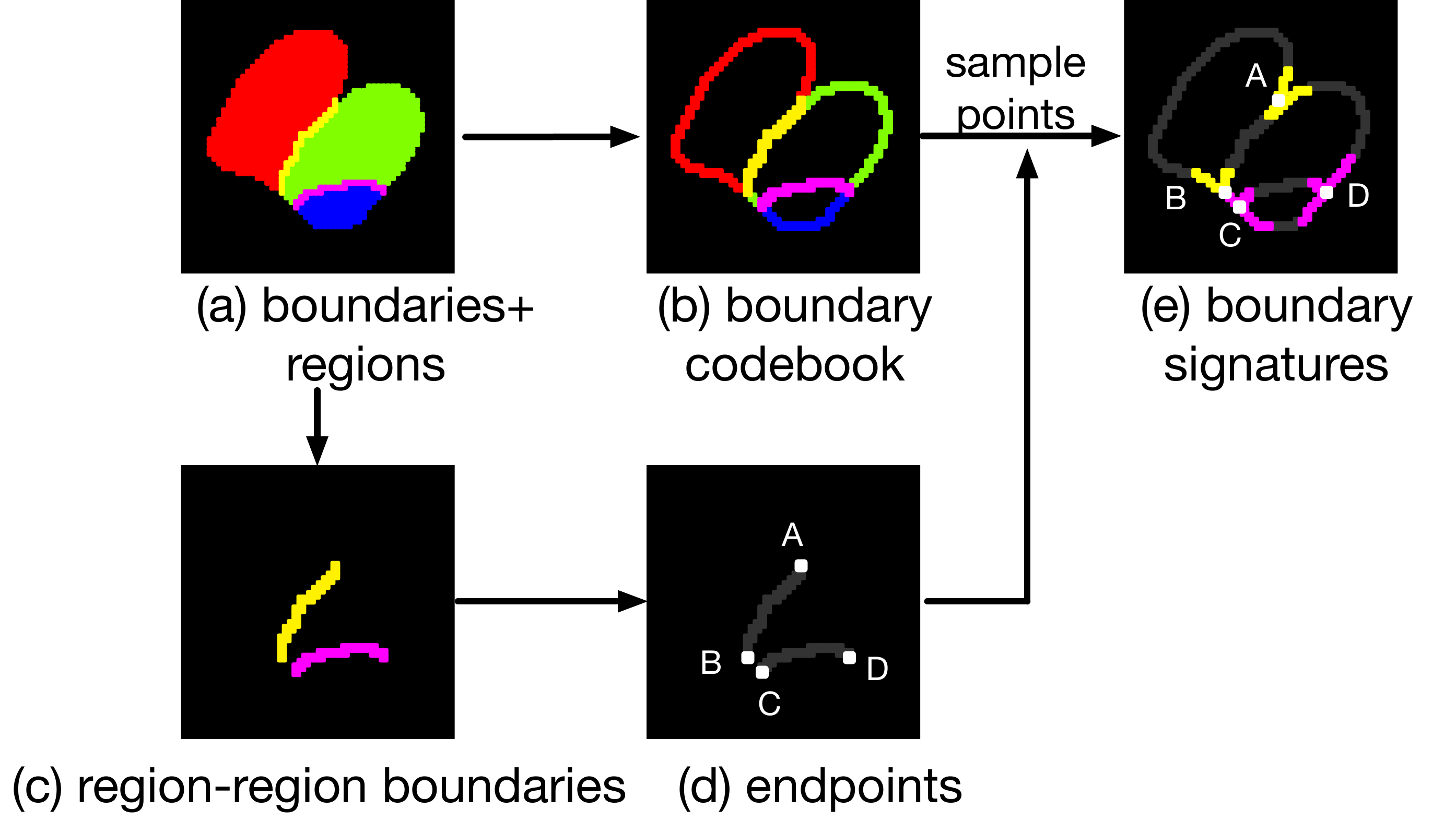}
\caption{The process of extracting 
boundary signatures. Given extracted boundaries and divided regions (a), we produce region-region boundaries (c), and obtain all possible boundaries, called the boundary codebook (b) which includes region-region boundaries and foreground-background boundaries. For each region-region boundary, we locate its two endpoints (d) and select a fork road around each endpoint including the corresponding boundary and two neighboring boundaries from the boundary codebook (b). We then sample pixels from the two fork roads and transform them into a binary mask to obtain a boundary signature (two boundary signatures are in (e)).
}
\label{fig:boundary_signature_extraction}
\end{figure}

The boundary signature extraction step extracts 
boundary-based features to represent some geometric properties of each boundary. As shown in Fig.~\ref{fig:boundary_signature_extraction} and Algorithm~\ref{algorithm:boundary_signature_extraction}, 
our strategy seeks to build a binary image for each boundary by sampling pixels around the two endpoints of the boundary. 

First, we apply the border following algorithm in~\cite{suzuki1985topological} to obtain foreground-background boundaries $\boldsymbol{\hat{B}}$. Each pixel of the foreground-background boundaries is indexed by its background and neighborhood regions. Hence, the foreground-background boundaries are divided into different segments (see the red, green, and blue lines in Fig.~\ref{fig:boundary_signature_extraction}(b)). The divided foreground-background boundaries, along with the region-region boundaries (e.g., see Fig.~\ref{fig:boundary_signature_extraction}(c)) obtained by Watershed~\cite{meyer1994topographic}, form the boundary set, called the boundary codebook. Next, for each region-region boundary, we create a weighted undirected graph, in which every pixel is a node and each neighboring pixel pair forms an edge with the distance between them as the edge weight. We apply the Floyd–Warshall algorithm~\cite{cormen2022introduction} to obtain the shortest path between every pixel pair in the graph. Then the endpoints of the boundary are the pixel pair with the largest-valued shortest 
path (e.g., see Fig.~\ref{fig:boundary_signature_extraction}(d)). We observe that there is a fork among the boundaries around each endpoint. Specifically, given two neighboring regions $\boldsymbol{R}_1, \boldsymbol{R}_2$ (e.g., the red and green regions in Fig.~\ref{fig:boundary_signature_extraction}(a)) with the corresponding region-region boundary $\boldsymbol{B}_1$ (the yellow boundary in Fig.~\ref{fig:boundary_signature_extraction}(a)), each endpoint of the region-region boundary is also attached to two other neighboring boundaries (the red and green boundaries in Fig.~\ref{fig:boundary_signature_extraction}(b)). We can query the neighboring boundaries in the boundary codebook to create the fork road around each endpoint. Finally, a boundary signature is obtained by sampling nearby pixels to the endpoints on both the fork roads and transforming them into a binary mask (e.g., see Fig.~\ref{fig:boundary_signature_extraction}(e)). Each boundary signature (as a binary mask) is associated with a true or false label (corresponding to that boundary, obtained in Section~\ref{section:boundary_label_assignment}). The boundaries and their labels will be input to the boundary classifier (Section~\ref{section:boundary_classification}) for training the model.

\subsection{Boundary Classification}
\label{section:boundary_classification}
With the extracted boundary signature and corresponding label for each boundary, we conduct binary boundary classification. Specifically, each boundary signature (an individual binary image, in the same format as MNIST~\cite{lecun1998mnist}) with its label is used as a training sample for the classifier. The associated boundary label (true or false) acts as ground truth. We utilize a Residual Network (ResNet-18)~\cite{he2016deep} as the classifier backbone, which predicts a boundary probability score $Prob(b)\in[0, 1]$ for a boundary $b$; a higher score indicates a higher probability of the boundary being a true boundary. Focal loss~\cite{lin2017focal} 
is used as the objective function.  

In our experiments, we observe a large distribution shift between the training probability map and the testing probability map (e.g., in the ratio of true boundaries and false boundaries and in the foreground-background boundary correctness). To deal with such distribution shift issues, we apply five-fold cross-validation on the training data, and use only the validation probability map as training data for our boundary classifier. In the inference stage, we apply the commonly-used threshold value of $0.5$ to identify true/false boundaries. Only true boundaries are preserved, and neighboring regions are merged after removing false boundaries. Fig.~\ref{fig:overview} presents an example of the inference process.

\subsection{Instance Temporal Setting}
\label{section:temporal}

\begin{figure*}[t]
\centering
\includegraphics[width=0.9\textwidth]{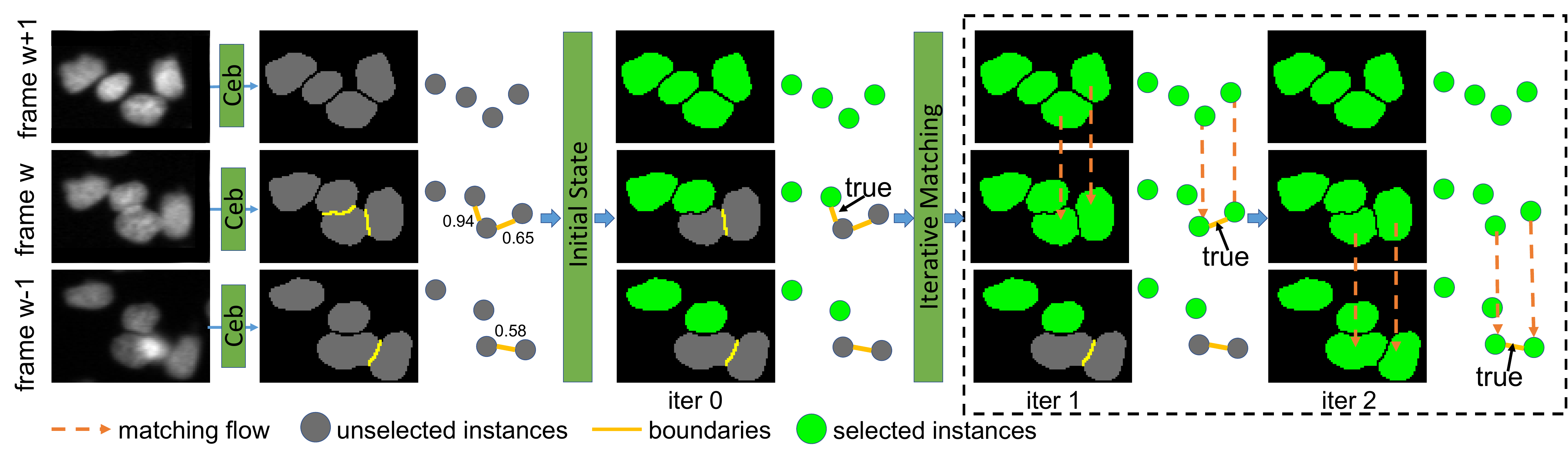}
\caption{An example illustrating our \textit{Ceb+Temporal} method applied to three consecutive frames, \(w-1\), \(w\), and \(w+1\). First, the \textit{Ceb} method generates regions, boundaries, and the associated boundary probability scores (shown as numerical values). Next, high-confidence boundaries (\textit{e.g.}, with scores $\geq 0.9$) are selected and the corresponding cell instances (attached only with high-confidence boundaries) are selected to form the initial state (marked in green at \(\mathrm{iter}\ 0\)).
In the 1st iteration (\(\mathrm{iter}\ 1\)), the selected instances from frame \(w+1\) are propagated to frame \(w\), allowing two previously unselected instances in frame \(w\) to be chosen. In the 2nd iteration (\(\mathrm{iter}\ 2\)), instances from frame \(w\) are propagated to frame \(w-1\), enabling the selection of two additional instances in that frame. All the selected instance candidates constitute the final instance segmentation results.
}
\label{fig:tem_overview}
\end{figure*}

In 2D temporal cell instance segmentation datasets, corresponding cell instances often exhibit both structural and distributional consistency across consecutive frames. For example, the same cell typically retains a stable position, size, and shape in a short time interval. These characteristics, known as temporal consistency, have been leveraged in prior research~\cite{liang2022h} to enhance instance segmentation. We develop an iterative matching and selection method, called \textbf{Ceb+temporal}, to incorporate temporal consistency. Fig.~\ref{fig:tem_overview} shows an example to illustrate our \textbf{Ceb+temporal} algorithm.
Given the regions and region-region boundaries (represented by a graph $G^w=(R^w, B^w)$, as in Section~\ref{section:boundary_generation}) for a frame $w$ and the boundary probability scores (generated by the boundary classifier, as in Section~\ref{section:boundary_classification}), our goal is to produce a set of final segmentation instances $U^w$ in frame $w$ from $G^w$. We first apply the boundary classifier to identify high-confidence false and true boundaries in $B^w$, which allow us to find easy-to-identify cell instances. These instances are selected to form an initial state $U_0^w$, and the corresponding selected nodes (with their adjacent edges) are removed from graph $G^w$ to obtain a reduced graph $\Bar{G}_0^w$ (\textbf{Creating an Initial State}). This reduced graph induces a new set of not-yet-selected possible instances $\Bar{I}_0^w$.
Next, the selected instances $U_t^{w-1}$ in frame $w-1$ and $U_t^{w+1}$ in frame $w+1$ are matched with the not-yet-selected instances $\Bar{I}_t^w$ in frame $w$ to select additional instances, forming an updated state $U_{t+1}^w$ of $w$, for $t=0, 1, \ldots, T-1$. The newly selected instances are removed from graph $\Bar{G}_t^w$, resulting in a reduced graph $\Bar{G}_{t+1}^w$. The not-yet-selected possible instances, $\Bar{I}_{t+1}^w$, are also determined. This matching and selection process is repeated for $T$ iterations to update the state (\textbf{Iterative Matching and Selection}).
Finally, the remaining possible instances in $\Bar{I}_{T}^w$ are selected using a \textbf{Final Selection} method. The instances selected with the Final Selection, $\mathcal{P}^w$, combined with the instances already in the state $U_T^w$, form the final segmentation instances $U^w$ in frame $w$. Below we present these steps in detail.

\subsubsection{Creating an Initial State}
For each frame $w$ in a video 
$X$, we represent its region-region boundaries and regions by an undirected graph $G^w=(R^w, B^w)$. We obtain the boundary probability score for each boundary $b \in B^w$ (see Section~\ref{section:boundary_classification}). 
The boundaries with probability scores lower than a threshold $\sigma_1$ are labeled as false, $B_{False}^w=\{b \ | \ Prob(b)<\sigma_1, b\in B^w\}$; the boundaries with probability scores larger than another threshold $\sigma_2$ are labeled as true, $B_{True}^w=\{b \ | \ Prob(b) > \sigma_2, b\in B^w \}$; the
remaining boundaries are marked as uncertain, $B_{UC}^w=\{b \ | \ \sigma_1  \leqslant Prob(b) \leqslant \sigma_2, b\in B^w \}$. 
The regions separated only by false boundaries are merged together, resulting in a new region set $\Bar{R}^w$. False boundaries and true boundaries are removed from the boundary set $B^w$, keeping only the uncertain boundaries. Consequently, the resulted regions $\Bar{R}^w$ and uncertain boundaries $B_{UC}^w$ form a new graph, $\Bar{G}^w=(\Bar{R}^w, B_{UC}^w)$. We then determine the
easy-to-identify instances as isolated nodes connecting to no other nodes in graph $\Bar{G}^w$ (i.e., their corresponding regions contain no uncertain boundaries). We select such regions as cell instances in the initial state: 
\begin{equation}
U_0^w=\{r \ | \ E(r)=\emptyset, r\in \Bar{R}^w \},
\label{eq:reliable_instance}
\end{equation}
where $E(r)$ is the set of edges adjacent to a node $r$ in $\Bar{G}^w$. The selected regions are then removed from graph $\Bar{G}^w$, resulting in a new graph $\Bar{G}^w_0=(\Bar{R}_0^w, B_{0}^w)$, with $\Bar{R}_0^w=\Bar{R}^w\setminus U_0^w$, $B_{0}^w=B_{UC}^w$. The not-yet-selected possible instances $\mathcal{\Bar{I}}^w_0$ contained in $\Bar{R}_0^w$ are obtained by enumerating all possible instances in $\Bar{G}^w_0$ (this process is described in Section~\ref{section:boundary_generation}).

\subsubsection{Iterative Matching and Selection} 
Given the selected instances in the state $U^w_t$
and not-yet-selected instances $\mathcal{\Bar{I}}^w_t$
of frame $w$ obtained in iteration $t$, we use the selected instances $U_t^{w-1}$ of frame $w-1$ and $U_t^{w+1}$ of frame $w+1$ to match with the not-yet-selected instances $\mathcal{\Bar{I}}_t^{w}$ of frame $w$ in iteration $t+1$. The matching and selection process is repeated for $T$ iterations. In each iteration, we perform three major substeps: selected-selected matching, selected-unselected matching, and state update, as follows.

\subsubsection*{Selected-selected Matching (SSM)}
For any two consecutive frames ($w, w+1$), we first perform a matching between all their selected instances (i.e., between $U_t^{w}$ and $U_t^{w+1}$). An instance in $U_t^{w+1}$ 
that is involved in any matched pair with an instance in $U_t^{w}$ is marked as ``occupied'', and thus should not be used to further match with any not-yet-selected instance in $\mathcal{\Bar{I}}_t^{w}$. 

We use the following matching model $\textbf{SSM}(U_t^{w+1}, U_t^{w})$ to compute an optimal matching between $U_t^{w+1}$ and $U_t^{w}$ ($\textbf{SSM}(U_t^{w-1}, U_t^{w})$ is for the $U_t^{w-1}$ and $U_t^{w}$ matching):
\begin{equation}\label{object_function_2}
  \textbf{SSM}(U_t^{w+1}, U_t^{w}) = \max_{f} \sum_{i \in U_t^{w+1}} \sum_{j \in U_t^{w}} M_{i,j} f_{i,j}
\end{equation}
\begin{equation}\label{constraint_2-1}
    \sum_{i \in U_t^{w+1}} f_{i,j} \leq 1, \forall j \in U_t^{w},
\end{equation}
\begin{equation}\label{constraint_2-2}
\sum_{j\in U_t^{w}} f_{i,j} \leq 1, \forall i \in U_t^{w+1},
\end{equation}
\begin{equation} \label{constraint_2-3}
f_{i,j} \in \{0,1\}, \forall i \in U_t^{w+1}, \forall j \in U_t^{w}.
\end{equation}

We utilize Intersection over Union (IoU) as the measure for the matching score $M_{i,j}$ between each pair of instances in the two frames. We solve this matching problem by integer linear programming (ILP) to obtain the optimal matching result (flows) $f_{i, j}$. Then, the matched instances in $U_t^{w+1}$ are removed to form the set $U_t^{w+1 \to w}$ of unmatched selected instances of frame $w+1$ with respect to frame $w$: $U_t^{w+1 \to w} = U_t^{w+1} \setminus \{i \ | \sum_{j \in U_t^{w}} f_{i,j} = 1, i \in U_t^{w+1} \}$.
Similarly, SSM is applied for the matching between frames $w-1$ and $w$: $U_t^{w-1 \to w} = U_t^{w-1} \setminus \{i \ |\sum_{j \in U_t^{w}}f_{i,j} = 1, i \in U_t^{w-1}  \}.$



\subsubsection*{Selected-Unselected Matching (SUM)}
Next, we use the unmatched but selected instances obtained with SSM (i.e., $U_t^{w+1 \to w}$ and $U_t^{w-1 \to w}$) to match with the not-yet-selected possible instances in $\mathcal{\Bar{I}}_t^{w}$ of frame $w$.
We define SUM from frame $w+1$ to frame $w$ as \textbf{SUM}$(U_t^{w+1 \to w}, \mathcal{\Bar{I}}_t^{w})$ (and that from frame $w-1$ to frame $w$ as \textbf{SUM}$(U_t^{w-1 \to w}, \mathcal{\Bar{I}}_t^{w})$):
\begin{equation}\label{object_function_3}
  \textbf{SUM}(U_t^{w+1\to w}, \mathcal{\Bar{I}}_t^{w}) = \max_{f} \sum_{i \in U_t^{w+1 \to w}} \sum_{j \in \mathcal{\Bar{I}}_t^{w}} M_{i,j} f_{i,j}
\end{equation}
\begin{equation}\label{equation:constraint_3-1}
    \sum_{j \in \mathcal{\Bar{I}}_t^{w}} f_{i,j} \leq 1, \forall i \in U_t^{w+1 \to w},
\end{equation}
\begin{equation}\label{equation:constraint_3-2}
\sum_{i\in U_t^{w+1 \to w}}\sum_{k \in K(j)} f_{i,k} \leq 1, \forall j \in \mathcal{\Bar{I}}_t^{w},
\end{equation}
\begin{equation} \label{equation:constraint_3-3}
    f_{i,j} \in \{0,1\}, \forall i \in U_t^{w+1 \to w}, \forall j \in \mathcal{\Bar{I}}_t^{w},
\end{equation}
where $M_{i,j}$ denotes the matching score for each pair of considered regions based on the Intersection over Union (IoU) measure.
Note that Eq.~(\ref{equation:constraint_3-2}) enforces that each region $r\in \Bar{R}_t^w$ can be matched at most once, where $K(r)$ denotes the set of all possible instance candidates that contain $r$.  

\subsubsection*{State Update}
Note that for each frame $w$ in a video $X$ except the first and last frames, $\mathcal{\Bar{I}}_t^{w}$ can receive matching results from both frame $w-1$ and frame $w+1$. This gives rise to two sets, $S_t^{w-1 \to w}$ and $S_t^{w+1 \to w}$, of matched instance candidates in $\mathcal{\Bar{I}}_t^{w}$. However, there may be inconsistencies between $S_t^{w-1 \to w}$ and $S_t^{w+1 \to w}$. 
Such inconsistencies in the matching results of $S_t^{w-1 \to w}$ and $S_t^{w+1 \to w}$ must be resolved. 
 
For this, we apply the following method. We call a maximal connected subgraph $C$ in an undirected graph $G$ (i.e., $C$ is not a subgraph of any larger connected subgraph in $G$) as a \textbf{component} of $G$~\cite{wiki:Component_graph_theory}.
Let $G_{cc}^w$ be the set of all the components in the graph $\Bar{G}_t^w$.
For each component $G_{cc_{i}}^w\in G_{cc}^w$, if the sum of the matching scores on $G_{cc_{i}}^w$ from frame $w-1$, $f_{G_{cc_{i}}^w}^{w-1 \to w}$, is larger than or equal to the sum of the matching scores on $G_{cc_{i}}^w$ from frame $w+1$, $f_{G_{cc_{i}}^w}^{w+1 \to w}$, then we use the matching results from $w-1$ as the matching results for $G_{cc_{i}}^w$; otherwise, we use the matching results from $w+1$ 
for $G_{cc_{i}}^w$. That is,
\begin{equation}
F_t^w =
\begin{cases} 
      \{f^{w+1 \to w}\} & w = 1, \\
      \{f^{w-1 \to w}\} & w = |X|, \\
      \{\max\{f_{G_{cc_{i}}^w}^{w+1 \to w}, f_{G_{cc_{i}}^w}^{w-1 \to w}\} \ | \ G_{cc_{i}}^w\in G_{cc}^w\} & \text{otherwise},
\end{cases}
\end{equation}
where $\{f^{w+1 \to w}\}$ and $\{f^{w-1 \to w}\}$ are the sets of matching scores from frames $w+1$ and $w-1$, respectively.
Let $\Delta U_t^w$ be the set of matched instance candidates in $\mathcal{\Bar{I}}_t^{w}$ corresponding to the matching scores in $F_t^w$. Then the state is updated as:
\begin{equation}
U_{t+1}^w = U_t^w \cup \Delta U_t^w.
\end{equation}
The graph is updated by removing the selected nodes and their adjacent edges: $\Bar{G}^w_{t+1}=(\Bar{R}_{t+1}^w, B_{t+1}^w)$, where $\Bar{R}_{t+1}^w = \Bar{R}_t^w \setminus \Delta U_{t+1}^w$, $B_{t+1}^w=B_t^w\setminus E(\Delta U_{t+1}^w)$, and $E(\Delta U_{t+1}^w)$ denotes the set of edges in $\Bar{G}_t^w$ adjacent to any node in $\Delta U_{t+1}^w$. The not-yet-selected instances, $\mathcal{\Bar{I}}^w_{t+1}$, are then obtained from $\Bar{G}^w_{t+1}$. 

\vspace{0.01in}
\subsubsection{Final Selection} 
After 
$T$ iterations of the above matching and selection process, some instance candidates in 
$\mathcal{\Bar{I}}_T^w$ may remain unselected, which may still be selected as instances. Thus, we apply a ``final selection'' process that assigns false or true labels to boundaries using a threshold value of $0.5$ on their boundary probability scores, obtaining a selected instance set $\mathcal{P}^w$.

The final instance set obtained in frame $w$ is the union of all the instances selected throughout, as:
\begin{equation} \label{equ:result2}
U^w = U_T^w \cup \mathcal{P}^w.
\end{equation}

$U=\{U^1,U^2, \ldots, U^{|X|}\}$ is taken as the final instance segmentation results of the input image video $X$.

\begin{table}[t]
\centering
\scriptsize 
\caption[dd]{Instance segmentation results on four video datasets. A {\bf bold} score marks the best performance on the corresponding dataset. An \underline{underline} score denotes the performance of the best foreground pixel clustering method with a specific 
backbone. ``--'' denotes that either KTH-SE is not applicable to the DIC-HeLa and PhC-U373 datasets, or CellViT, UN-SAM, and GAC do not perform well on the corresponding datasets.}
\setlength{\tabcolsep}{1mm}
{

\begin{tabular}{  c| c |c c |c c |cc|cc }\hline

 \multicolumn{2}{c|}{}&
 \multicolumn{2}{c|}{DIC-HeLa} &
 \multicolumn{2}{c|}{Fluo-HeLa} &
 \multicolumn{2}{c|}{PhC-U373} &
 \multicolumn{2}{c}{Fluo-SIM+} \\ \cline{3-10}
 \multicolumn{2}{c|}{} 
 &  F1   & AJI   
 & F1   & AJI  
 & F1   & AJI 
 & F1   & AJI 
  \\ \hline
 \multicolumn{2}{c|}{KTH-SE~\cite{ulman2017objective}} 
  & -- & --
 & 96.3 & 90.0 
 & -- & -- 
 & 97.9 & \textbf{87.8}  \\ 
  \multicolumn{2}{c|}{CellPose~\cite{pachitariu2022cellpose}} 
  & 95.4 & 84.4
  &96.2 &91.0 
  & 93.3 & 89.4 
  & 96.8 & 78.9 \\
 \multicolumn{2}{c|}{Mask R-CNN\cite{he2017mask}} 
 & 93.7 & 71.6
 &90.7 & 77.0  
 &67.6  & 62.5 
 & 90.4 & 76.9   \\
  \multicolumn{2}{c|}{StarDist~\cite{schmidt2018cell}
  } 
   & 96.4 & 80.1
  &\textbf{96.7} &91.1
  & 93.4 &82.2
   & 96.2 & 79.1 \\
  \multicolumn{2}{c|}{KIT-Sch-GE~\cite{scherr2021improving}} 
   & 77.3 & 58.1
  & 96.6 & 92.6
  &93.2 & 79.6 
  &95.7 & 81.5 \\
  \multicolumn{2}{c|}{nnU-Net~\cite{isensee2021nnu}}
 & 91.1 & 78.7
  & 93.1 & 83.7 
  &92.3 & 86.3 
  & 97.9  & 83.9  \\
  \multicolumn{2}{c|}{InstanSeg~\cite{goldsborough2024novel}} & 92.8 & 77.4 & 96.3 & 92.2 & 93.7 & 87.9 &
  94.6 & 80.3 \\
  \multicolumn{2}{c|}{CellViT~\cite{horst2024cellvit}}
  & -- & --
  & 88.0 & 84.6
  & 66.8 & 70.6 &
81.6 & 70.4 \\
  \multicolumn{2}{c|}{FCIS~\cite{zhang2025fourcolor}}
  & 44.6 & 32.3
  & 88.6 & 80.1
  & 84.3 & 75.8 
  & 90.4 & 71.7 \\
  \multicolumn{2}{c|}{UN-SAM~\cite{chen2025sam}}
  & -- & -- 
  & 88.5 & 82.9 
  & 89.0 & 87.4 &
  98.1 & 85.5 \\
  \multicolumn{2}{c|}{CelloType~\cite{pang2025cellotype}}
  & 94.2 & 82.6 
  & 51.8 & 31.4 
  & 85.7 & 70.2
  & 89.9 & 71.2 \\
  \hline \hline
\multirow{11}{*}{\tabincell{c}{
U-Net\\~\cite{ronneberger2015u}}
} 
& 0.5-Th
& 73.5 & 57.6
& 93.4 & 84.8
& 92.4 & 88.5
& 92.9 & 75.5  \\
 & Otsu's 
  & 72.8 & 56.9
 & 93.2 & 84.4 
 & 91.3 & 89.0 
 & 92.6 & 74.9   \\
 & DenseCRF 
 & 70.6  & 55.4
 & 91.4 & 80.0  
 & 92.5 & 89.0 
 & 91.6 & 73.1 \\
 & MaxValue 
 & 73.3 & 57.2
 & 93.3 & 84.5 
 & 91.6 & 88.9 
 & 91.4 & 72.1  \\
 & H-EMD
& 87.2 & 72.0
& 96.4 & 92.4 
& 93.5 & 89.4 &97.6 & 84.3  \\ \cline{2-10}
& GAC & -- & -- & 82.6 & 61.4 & 89.4 & 79.8 & 89.1 & 73.2  \\
& ACWE & 61.6 & 51.4 & 88.5 & 72.9 & 93.0 & 87.6 & 84.7 & 64.7  \\
 & Watershed
 & 93.0 & 83.9 
 & 95.7 & 91.5
 & 91.6 & 89.4
 & 97.6 & 83.8 \\
 & Ceb w/o cls
& 93.5 & 83.6
& 95.7 & 91.3
& 80.0 & 73.5
& 97.0 & 83.8  \\
& Ceb
& 96.1 & 84.6
& 96.4 & 92.4 
& 93.2 & 88.7
& 97.7 & 84.5  \\
& Ceb+temporal
& \underline{97.1} & \underline{85.0}
& \underline{96.6} & \underline{\textbf{92.9}}
& \underline{93.5} & \underline{89.5}
& \underline{97.8} & \underline{84.7} 
\\ \hline

\multirow{11}{*}{\tabincell{c}{DCAN\\~\cite{chen2016dcan}}} 
&0.5-Th 
& 62.9 & 50.0
& 92.3 & 82.8 
& 89.6 & 87.6 
& 92.6 & 76.2  \\
& Otsu's 
& 64.5 & 49.5
& 92.1 & 82.5  
& 89.4 & 88.1 
& 92.5 & 75.6  \\
 & DenseCRF 
  & 62.6 & 45.8
 & 91.2 & 80.0
 & 91.0  & 88.1 
 & 91.7 & 74.0  \\
 & MaxValue
  & 62.5 & 47.7
 & 92.6 & 83.5 
 & 89.0 & 88.0 
 & 91.5 & 73.3 \\
 & H-EMD
& 82.2 & 66.6
& 96.1 & 91.6  
& 93.3 & 88.7 
& 98.0 & 85.2  \\ \cline{2-10}
 & GAC & -- & -- & 82.2 & 61.4 & 89.7 & 80.0 & 89.0 & 72.1  \\
& ACWE & 42.5 & 31.8 & 88.4 & 72.9 & 92.9 & 86.9 & 84.6 & 63.6  \\
 & Watershed 
 & 87.9 & 80.1
 & 95.5 & 91.4  
 & 91.2 & 88.4 
 & 97.7 & 83.7 \\
& Ceb w/o cls
& 88.6 & 81.1
& 96.0 & 92.0 
& 93.3 & 89.3
& 98.0 & 84.4  \\
& Ceb 
& 91.3 & 81.3
& 96.3 & 92.2  
& 94.0 & \textbf{\underline{89.6}}
& \textbf{\underline{98.3}} & 85.5  \\
& Ceb+temporal
& \underline{93.0} & \underline{83.9}
& \underline{96.6} & \underline{92.4} 
& \underline{\textbf{94.1}} & 89.4 
& \underline{\textbf{98.3}} & \underline{85.7}  \\ \hline

\multirow{11}{*}{\tabincell{c}{
Res2Net\\~\cite{gao2019res2net}}} & 0.5-Th 
& 59.7 & 45.5
& 91.1 & 80.8      
& 87.9 & 86.4  
& 85.4 & 56.4   \\
& Otsu's
& 59.6 & 45.0
& 91.0 & 80.6    
& 87.4 & 86.3  
& 85.0 & 55.1   \\
& DenseCRF 
& 57.2 & 43.2
& 90.9 & 79.7   
& 90.8 & 86.3  
& 83.4 & 53.5  \\
& MaxValue
& 56.3 & 41.5
& 90.9 & 80.6      
& 87.0 & 86.7  
& 82.6 & 51.4   \\
& H-EMD
& 88.7 & 77.3 
& 95.8 & 90.6
& 92.2 & 86.9
& 96.2 & 73.3 \\ \cline{2-10}
& GAC & -- & -- & 82.2 & 61.1 & 88.5 & 79.1 & 78.1 & 56.1  \\
& ACWE & 47.4 & 33.6 & 88.4 & 73.0 & 91.4 & 85.4 & 71.7 & 40.2  \\
& Watershed
& 92.8 & 83.6
& 95.5 & 91.5  
& 84.6 & 83.8  
& 96.2 & 73.2   \\
& Ceb w/o cls 
& 84.7 & 77.4
& 95.2 & 91.1  
& 90.8 & 85.8  
& 95.7 & 73.3  \\
& Ceb  
& 93.5 & 83.7
& 96.0 & 91.6   
& 92.4 & 86.6   
& \underline{96.5} & \underline{73.9} \\
& Ceb+temporal 
& \underline{\textbf{97.6}} & \underline{\textbf{86.5}}
& \underline{96.4} & \underline{92.2} 
& \underline{93.6} & \underline{87.9}
& 96.3 & 73.6  \\ \hline

\end{tabular}
\label{tab:public_video}
}
\end{table}

\section{Experiments}

\begin{table}[t]
\centering
\scriptsize
\caption[dd]{Instance segmentation results on two 2D datasets. 
{A {\bf bold} score marks the best performance on the corresponding dataset. An \underline{underline} score denotes the performance of the best foreground pixel clustering method with a specific 
backbone. ``--'' denotes that either KIT-Sch-GE and nnU-Net are not applicable to the corresponding dataset, or GAC does not perform well on the corresponding dataset.}
}
\setlength{\tabcolsep}{1mm}
{

\begin{tabular}{ c| c |c c c|c c c}\hline

& &  \multicolumn{3}{c|}{BBBC039} &
 \multicolumn{3}{c}{TissueNet} \\ \cline{2-8}
&  & F1 $(\%)$  & AJI $(\%)$ & AP $(\%)$ & F1 $(\%)$ & AJI $(\%)$  & AP $(\%)$ \\ \hline
& CellPose
  & 95.0 & 89.2 & 90.0 
  & 94.3 & 82.1 & 88.0 \\
& Mask R-CNN
 & 94.3 & 86.0 & 88.8 
 & 94.0 & 82.3 & 87.0 \\
& StarDist
  & 94.5 & 88.3 & 89.5 
  & 92.9 & 82.2 & 86.0 \\
& KIT-Sch-GE
  & 89.6 & 79.0 & 80.5 
  & -- & -- & -- \\
 & nnU-Net
  & 94.2 & 85.8 & 89.2
  & -- & -- & -- \\
  & InstanSeg
  & 95.0 & 89.4 & 90.4
  & 94.4 & 84.3 & 88.6 \\ 
  & CellViT
  & 85.2 & 80.5 & 75.6
  & 93.5 & 84.2 & 87.2 \\
  & FCIS
  & 88.2 & 81.0 & 77.0
  & 83.4 & 75.2 & 69.5 \\
    & UN-SAM
  & 89.7 & 76.3 & 79.6
  & 75.7 & 60.0 & 60.3 \\
  & CelloType
  & 80.9 & 61.1 & 66.8
  & 90.2 & 69.8 & 80.2 \\
  \hline \hline
 \multirow{9}{*}{\tabincell{c}{
U-Net}}
& 0.5
& 88.2 & 75.9 & 76.8 
& 66.9 & 40.4 & 50.3 \\
 & Otsu's & 87.9 & 75.5 & 76.3  
 & 64.4 & 37.2 & 47.5 \\
 & DenseCRF & 86.8 & 73.7 & 74.3 
 & 57.9 & 31.1 & 41.1 \\
 & MaxValue
& 88.0 & 75.1 & 75.2 
& 67.1 & 40.9 & 51.1 \\ \cline{2-8}
& GAC & 76.6 & 59.3 & 58.7 & -- & -- & -- \\
& ACWE & 79.3 & 64.0 & 62.9 & 42.6 & 32.6 & 30.2 \\
& Watershed 
 & 94.5 & 87.8 & 89.2 
 & 91.4 & 82.7 & 82.4 \\
 & Ceb w/o cls
  & 93.5 & 88.4 & 87.1
  & 91.2 & 80.0 & 82.0 \\ 
 & Ceb (ours) 
  & \underline{95.0} & \underline{89.5} & \underline{90.4} 
  & \underline{94.5} & \underline{83.2} & \underline{88.5} \\\hline

\multirow{9}{*}{\tabincell{c}{
DCAN}} & 0.5-Th & 87.2 & 73.8 & 74.9
& 67.0 & 40.3 & 50.8 \\
& Otsu's & 87.1 & 73.4 & 74.9
& 64.3 & 37.0 & 47.8 \\
& DenseCRF & 86.1 & 72.0 & 73.0 
& 57.1 & 30.3 & 40.6 \\
& MaxValue
& 87.4 & 73.8 & 75.4 
& 67.2 & 40.5 & 50.9 \\ \cline{2-8}
& GAC & 76.6 & 59.7 & 58.6 & -- & -- & --  \\
& ACWE & 79.7 & 65.3 & 63.7 & 40.1 & 31.0 & 29.8 \\
& Watershed & 94.6 & 87.4 & 88.6 
& 92.0 & 84.1 & 83.5 \\
& Ceb w/o cls & 94.8 & 89.4 & 90.0 
& 92.3 & 81.8 & 84.0 \\
& Ceb (ours) & \textbf{\underline{95.2}} & \textbf{\underline{89.6}} & \textbf{\underline{90.6}} 
& \underline{94.7} & \underline{84.3} & \underline{88.8} \\ \hline

  \multirow{9}{*}{\tabincell{c}{
Res2Net}} & 0.5-Th & 87.0 & 74.0 & 74.6
& 63.8 & 36.5 & 47.1 \\
& Otsu's & 86.9 & 73.8 & 74.4 
& 62.4 & 35.0 & 45.5  \\
& DenseCRF & 86.0 & 72.3 & 72.9
& 55.9 & 28.9 & 39.3  \\
& MaxValue
& 86.7 & 73.4 & 74.3 
& 63.1 & 36.1 & 46.4 \\ \cline{2-8}
& GAC & 76.0 & 58.6 & 57.9 & -- & -- & -- \\
& ACWE & 79.3 & 64.3 & 63.0 & 39.8 & 29.8 & 28.3 \\
& Watershed & 92.9 & 86.2 & 85.7 
& 91.8 & 84.6 & 83.1 \\
& Ceb w/o cls & 90.0 & 84.3 & 80.9 
& 93.1 & 83.1 & 85.6 \\
& Ceb (ours) & \underline{94.1} & \underline{88.0} & \underline{88.8}
& \textbf{\underline{95.3}} & \textbf{\underline{85.3}} & \textbf{\underline{90.0}} \\ \hline

\end{tabular}
\label{tab:2D}
}
\end{table}

The experiments evaluate our Ceb approach for cell instance segmentation by applying Ceb on top of probability maps produced by three representative semantic segmentation models. 

\subsection{Datasets}
We evaluate our framework on four cell video datasets from the Cell Tracking Challenge~\cite{ulman2017objective} (Fluo-N2DL-HeLa, Fluo-N2DH-SIM+, PhC-C2DH-U373, and DIC-C2DH-HeLa) 
and two 2D cell datasets (BBBC039~\cite{caicedo2019evaluation} and TissueNet~\cite{greenwald2022whole}). Each cell video dataset contains two sequences with annotated labels. We use one sequence
as training data and the other sequence
as test data. Specifically, Fluo-N2DH-SIM+ uses the second video for training and the first video for testing, and vice versa for all the other datasets.
The BBBC039 dataset contains $100$ training and $50$ test images of nuclei of U2OS cells collected with fluorescent microscopy. The TissueNet
dataset contains $2,601$ training and $1,249$ test images of six different tissue types collected with fluorescent microscopy; each image has manual segmentations of cells and nuclei. We use the nuclei segmentation labels in this study. 

\begin{figure*}[t]
\centering
\includegraphics[width=\textwidth]{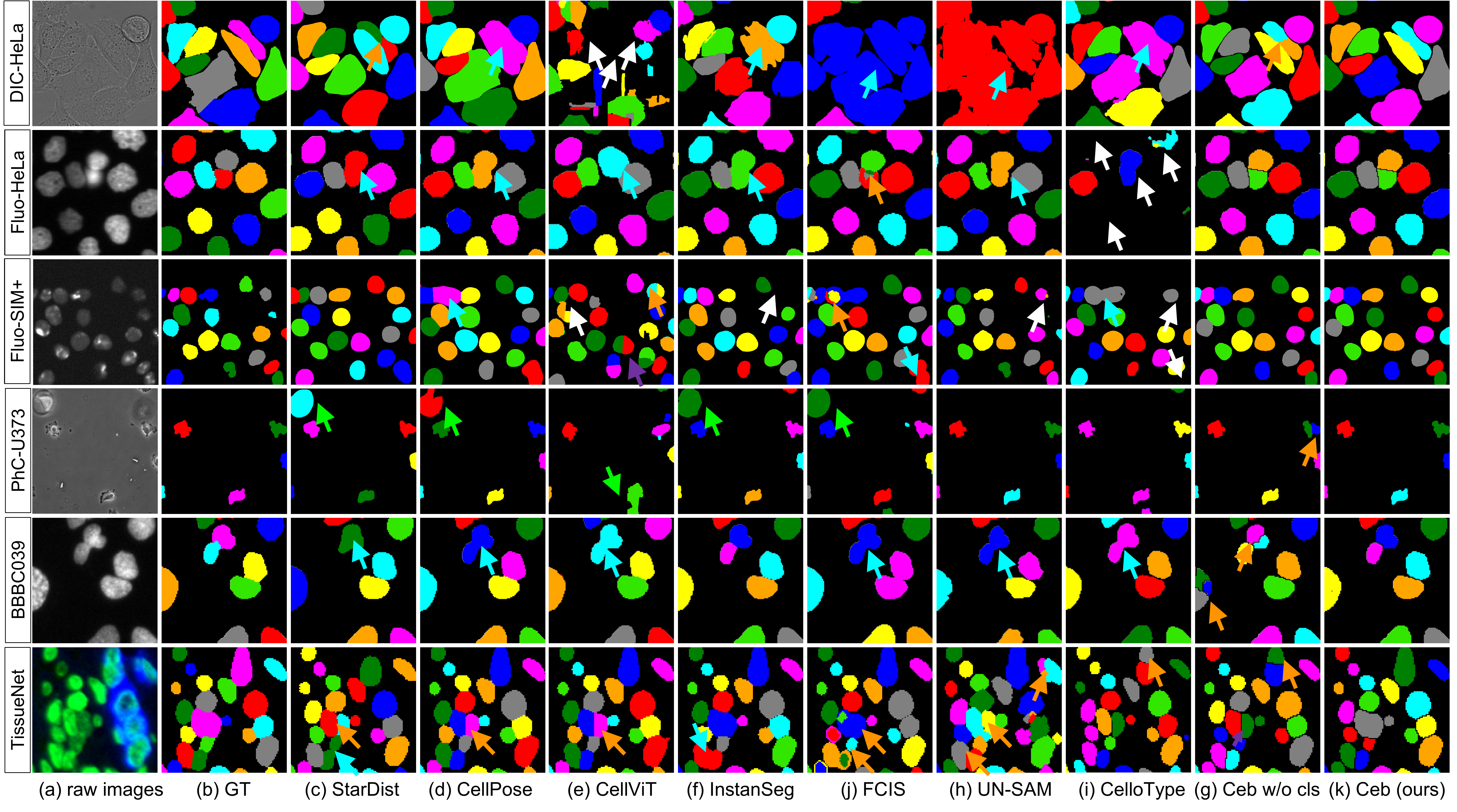}
\caption{Visual examples of cell instance segmentation results. GT denotes ground truth.
Orange arrows point to some over-segmentation errors corrected by Ceb. 
Cyan arrows point to 
some under-segmentation errors corrected by Ceb. White arrows point to false negative errors corrected by Ceb. Green arrows point to false positive errors corrected by Ceb.}
\label{fig:experiment_visualization}
\end{figure*}

\subsection{Compared Methods}
We compare our Ceb framework with two major categories of the known cell instance segmentation methods. 
\subsubsection{Semantic Segmentation}
These methods were built on different image encoders (\textit{i.e.}, backbones) with semantic segmentation objectives (\textit{i.e.}, pixel-wise classification) and different foreground pixel clustering strategies. Specifically, we use three widely-used backbones: U-Net~\cite{ronneberger2015u}, DCAN~\cite{chen2016dcan}, and Res2Net~\cite{gao2019res2net}. We then apply $8$ different foreground pixel clustering methods, which are categorized into the following two types.

\noindent
\textbf{Boundary-generation based methods:}
\begin{itemize}
\item 
Geodesic Active Contour (GAC)~\cite{caselles1997geodesic}: It evolves a contour toward object boundaries by optimizing an energy function that depends on image gradients.
\item Active Contours without Edges (ACWE)~\cite{chan2001active}: It segments images by optimizing an energy function based on the differences in intensity between inside and outside regions of the contour.
\item Watershed~\cite{meyer1994topographic}: It takes probability maps as input to generate cell segmentation results. To mitigate over-segmentation issues, we smooth the probability maps following previous work~\cite{liang2022h}.  
\end{itemize}

\textbf{The other methods:}
\begin{itemize}
    \item 0.5 thresholding (0.5-Th)~\cite{zhou2019cia}: The probability maps are binarized using the threshold value of 0.50, and connected components are computed as the final instances.  
    \item Otsu's~\cite{otsu1979threshold}: Otsu's algorithm is a global thresholding method that automatically determines the optimal threshold by maximizing the between-class variance of pixel intensities in a probability map.
    \item DenseCRF~\cite{kamnitsas2017efficient}: Both the raw images and their probability maps are fed to the DenseCRF model. Pixels with similar features (\textit{e.g.}, color and probability) are assigned to the same semantic class (\textit{e.g.,} foreground or background).
    \item MaxValue~\cite{liang2022h}:  Each pixel is assigned to one of three classes (background, boundary, and foreground) with the maximum probability.
    \item H-EMD~\cite{liang2022h}:  Given probability maps across a sequence of cell images (\textit{e.g.}, in a video), instance candidates are first generated from the probability maps, and temporal consistency is leveraged to select an optimal subset of instance candidates as the final instance segmentation results.
     \end{itemize}

\subsubsection{SOTA Cell Instance Segmentation Methods}
Note that some cell instance segmentation methods do not necessarily produce semantic segmentation probability maps. We consider the following SOTA methods: 

\begin{itemize}
\item 
KTH-SE~\cite{ulman2017objective}: It uses a bandpass filtering based segmentation algorithm~\cite{mavska2014benchmark} to segment cells and applies the Viterbi tracking algorithm~\cite{magnusson2014global} to correct potential segmentation errors.  
\item
CellPose~\cite{pachitariu2022cellpose}: It predicts a spatial gradient vector field pointing from pixels within each cell toward its centroid, and uses this field to reconstruct individual cell instances.  
\item
StarDist~\cite{schmidt2018cell}: It predicts a set of radial distances (32 in the experiments) along fixed angles from each pixel to the object boundaries. 
\item
KIT-Sch-GE~\cite{scherr2021improving}: It predicts cell distance maps, to which the watershed algorithm is applied to generate the final instance segmentation results. 
\item
nnU-Net~\cite{isensee2021nnu}: It is an automatic self-configured U-Net-based model including pre-processing, network architecture, training, and post-processing for biomedical image segmentation. 
\item
Mask R-CNN~\cite{he2017mask}: It first generates instance proposals and then predicts segmentation masks within each proposed region.
\item
InstanSeg~\cite{goldsborough2024instanseg}: It predicts seed points that represent instance centers and learns pixel-wise embeddings; pixels are then clustered into instances based on their similarity to the seed embeddings.
\item
CellViT~\cite{horst2024cellvit}: It employs a distance-map based approach and leverages Vision Transformer (ViT)~\cite{dosovitskiy2020image} encoders. 
\item FCIS~\cite{zhang2025fourcolor}: It encodes foreground pixels by assigning identical values to pixels within the same instance while ensuring neighboring instances receiving different values. 
\item 
UN-SAM~\cite{chen2025sam}: It fine-tunes the Segment Anything Model (SAM)~\cite{kirillov2023segment} to adapt it for cell instance segmentation. 
\item CelloType~\cite{pang2025cellotype}: It employs a Transformer-based detector (DINO)~\cite{caron2021emerging} to generate cell proposals, followed by segmentation of cell instances.  
\end{itemize}

Note that for KTH-SE, we are able to evaluate the method on the Fluo-N2DL-HeLa and Fluo-N2DH-SIM+ datasets, for which KTH-SE was originally designed. For H-EMD, it was designed for cell instance segmentation in videos, and is not directly suitable for general 2D cell instance segmentation tasks. We use the publicly released codes of the corresponding methods, and train their models in the same train/test split settings. 

\subsection{Evaluation Metrics}
We utilize two widely-used cell instance segmentation evaluation metrics: F1-score~\cite{zhou2019cia} and Average Jaccard Index (AJI)~\cite{zhou2019cia} for video datasets. For 2D cell datasets, we additionally use the Average Precision (AP) metric~\cite{stringer2021cellpose}, a standard measure for these datasets.

\subsection{Implementation Details}

\begin{table*}[t]
\centering
\caption[dd]{Summary of implementation details of the DL backbones that we use.}
\setlength{\tabcolsep}{1mm}
{
\begin{tabular}{ c c c c c c c c c c }
\hline
Functionality & Model & Framework & Initialization & Optimizer & Learning Rate & Augmentation & Batch & Loss & Input size \\ \hline
Segmentation & U-Net & TensorFlow & Gaussian & Adam & 5e-4 & Rotate + Flip & 8 & Cross-Entropy & 192  \\
Segmentation & DCAN & TensorFlow & Gaussian & Adam & 5e-4 & Rotate + Flip & 8 & Cross-Entropy & 192 \\
Segmentation & Res2Net & PyTorch & Gaussian & Adam & 5e-4 & Rotate + Flip & 8 & Cross-Entropy & 192  \\ \hline
Boundary Classifier & ResNet-18 & PyTorch & Pre-trained & SGD & 1e-3 & Rotate + Crop + Flip & 8 & Focal Loss & 224 \\
\hline

\end{tabular}
\label{tab:DL_details}
}
\end{table*}

Table~\ref{tab:DL_details} summarizes the implementation details of the three semantic segmentation backbones we use (U-Net, DCAN, and Res2Net) and the boundary classifier we use (ResNet-18). All the training and inference procedures are performed on an NVIDIA Tesla V100 GPU with 32 GB of memory. The semantic segmentation task is formulated as a three-class classification problem consisting of foreground, boundary, and background regions. The boundary class is generated by applying a three-pixel dilation to the ground truth masks.

Our Ceb framework involves a handful of hyperparameters. In the Seed Generation stage, to filter out noisy instance candidates, we use only threshold values $>0.50$ to generate instance candidate forests (ICFs). In the Boundary Classification stage, the threshold value for the boundary classifier is set to 0.50. The Integral Linear Programming (ILP) problems are solved using the GLPK solver\footnote{https://www.gnu.org/software/glpk/}, in which the matching scores are defined by the Intersection over Union (IoU) metric. For the temporal experimental setting, the hyperparameter $T$, representing the number of iterations, is set to 10.

\subsection{Results}


\begin{figure*}[t]
    \centering
    \begin{subfigure}{0.35\textwidth}
        \centering
        \includegraphics[width=\textwidth]{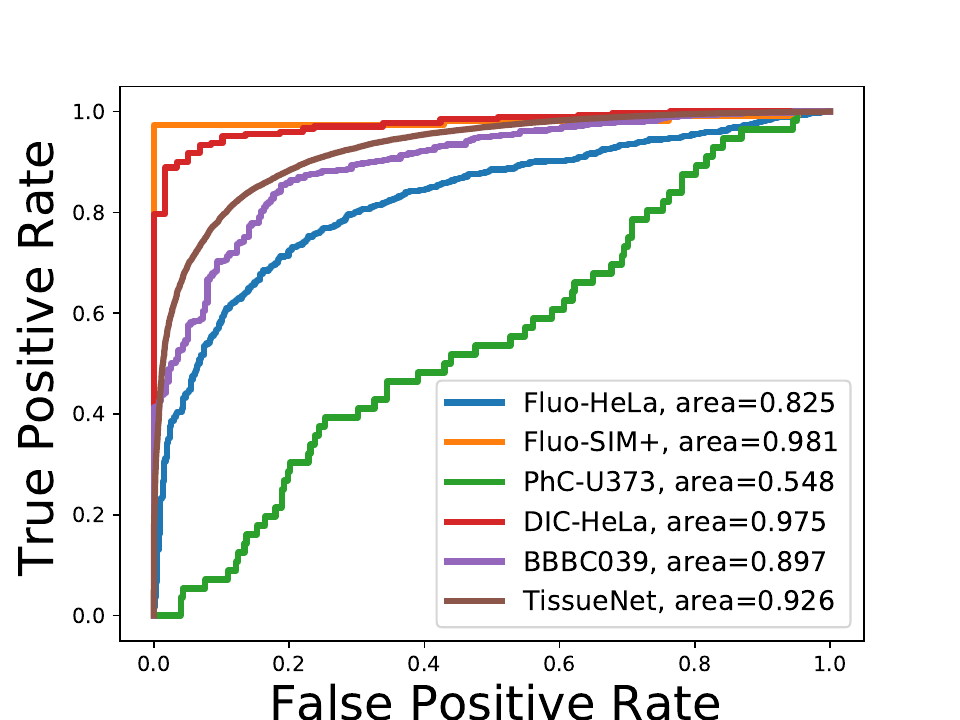}
        \caption{}
        \label{fig:ROC}
    \end{subfigure}
    \begin{subfigure}{0.4\textwidth}
        \centering
        \includegraphics[width=\textwidth]{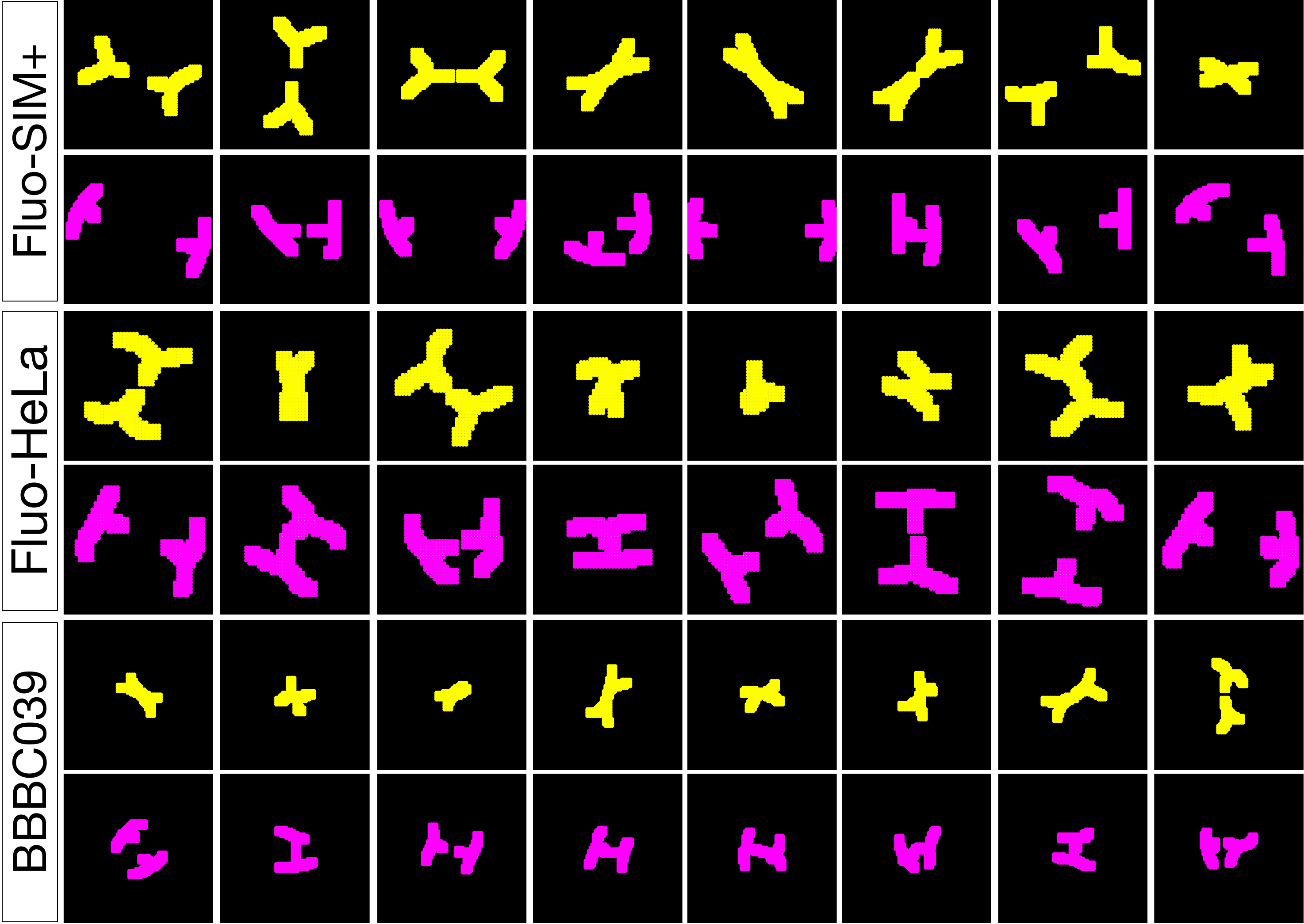}
        \caption{}
    \label{fig:sigature_visualization}
    \end{subfigure}
    \caption{(a) Receiver operating characteristic (ROC) curves of the boundary classifier on the six datasets based on a U-Net backbone. (b) True and false boundary signature samples in three different datasets.
\protect\tikz \protect\fill[magenta] (1ex,1ex) rectangle (0.6, 0.3);: false samples;  
\protect\tikz \protect\fill[yellow] (1ex,1ex) rectangle (0.6, 0.3);: true samples.}
    \label{fig:side_by_side}
\end{figure*}

\begin{figure*}[t]
\centering
\includegraphics[width=0.9\textwidth]{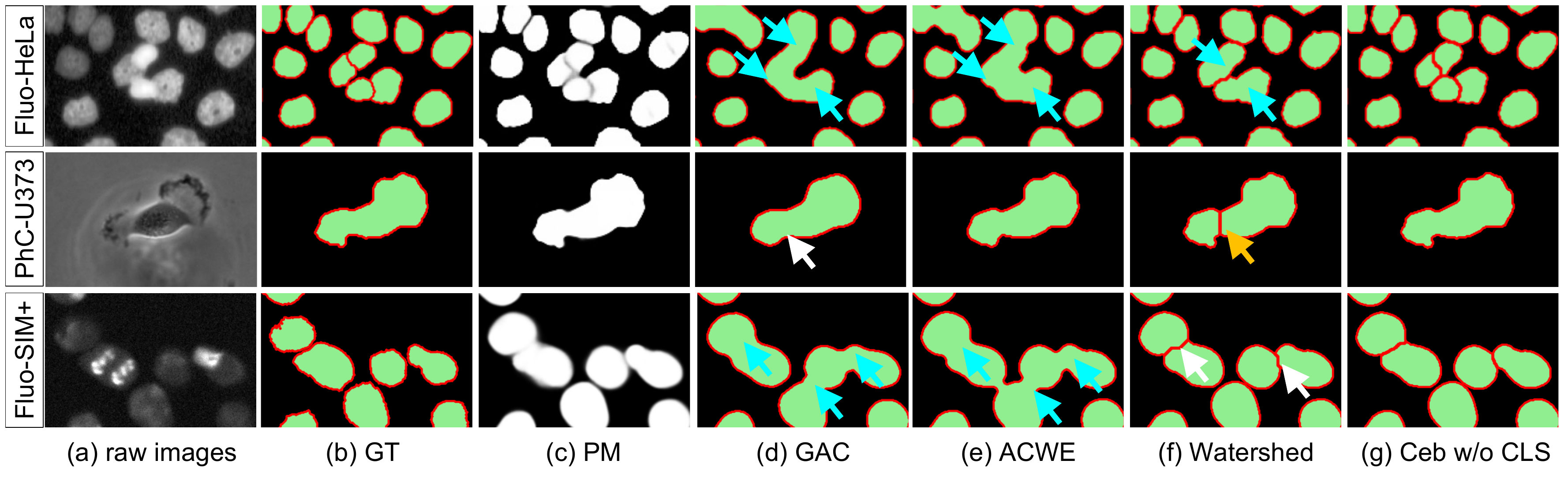}
\caption{Visual comparisons of different boundary generation methods. GT denotes ground truth; PM denotes probability map; GAC denotes Geodesic Active Contours; ACWE denotes Active Contours without Edges. Orange arrows indicate over-segmentation errors; cyan arrows indicate under-segmentation errors; white arrows indicate visual boundary errors.}

\label{fig:boundary_visualization}
\end{figure*}


\subsubsection{Cell Video Dataset Results}
Table~\ref{tab:public_video} shows the instance segmentation results on the four cell video datasets. 
We consider three settings for our method: \textit{Ceb} represents our full model, and \textit{Ceb w/o cls} represents our method without boundary classification (\textit{cls} is short for classification). In other words, \textit{Ceb w/o cls} takes all the extracted region-region boundaries as true boundaries. 
\textit{Ceb + Temporal} represents our method in the temporal setting, which incorporates instance temporal consistency to determine the final instance results.
Based on the results in Table~\ref{tab:public_video}, we draw three conclusions. First, compared to \textit{Ceb w/o cls}, \textit{Ceb} shows consistent improvements across all the datasets with all the semantic segmentation backbone models. Second, compared to other foreground pixel clustering methods that extract instances from probability maps, \textit{Ceb} yields the best performances across all the datasets. 
Compared to the best foreground pixel clustering performances, \textit{Ceb} improves the F1 score by $3.4\%$ and AJI by $1.0\%$ on the DIC-C2DH-HeLa dataset with the DCAN backbone. We also notice that even though H-EMD already effectively utilizes temporal information in videos, \textit{Ceb} still outperforms H-EMD in most of the cases without using any temporal information.  
Third, \textit{Ceb + Temporal} further improves performance compared to \textit{Ceb} by incorporating instance temporal consistency. 
\textit{Ceb + Temporal} outperforms the
SOTA instance segmentation methods in most the metrics.
We observe that CellViT does not yield good performance on the video datasets, due to the limited training data available, 
as ViT-based architectures typically require large-scale training data sets. 
However, this limitation is common in cell datasets, where acquiring extensive training data can be both time-consuming and expensive.

\noindent
\subsubsection{2D Dataset Results}
On the two general 2D cell datasets, BBBC039 and TissueNet, 
Table \ref{tab:2D} shows the results. One can see that \textit{Ceb}
consistently boosts instance segmentation performances. Compared to the other foreground pixel clustering methods, \textit{Ceb} yields the best scores of all the backbones. On the BBBC039 dataset, \textit{Ceb} improves F1 by $1.2\%$, AJI by $1.8\%$, and AP by $3.1\%$ with a Res2Net backbone compared to the best results of the known foreground pixel clustering methods. On the TissueNet, \textit{Ceb} improves F1 by $3.5\%$, AJI by $0.7\%$, and AP by $6.9\%$ with a Res2Net backbone compared to the best results of the known foreground pixel clustering methods. Compared to the SOTA instance segmentation methods, \textit{Ceb} still achieves the best results across all the metrics. 

Fig.~\ref{fig:experiment_visualization} shows some visual results
of cell instance segmentation on four datasets. One can see that for instances
that are over-segmented or under-segmented by other methods, Ceb can
attain correct instance-level segmentation results. 

\subsubsection{Cell Tracking Evaluation on Video Datasets} 
To evaluate the effectiveness of incorporating temporal consistency by our Ceb+temporal method, we conduct additional tracking experiments. Specifically, we apply the EMD-based tracking model in~\cite{chen2016segmentation} to the cell instance segmentation results obtained by Ceb and Ceb+temporal (both using the U-Net backbone). We use the Tracking Accuracy (TRA) metric~\cite{ulman2017objective} to measure how accurately cell instances are tracked across frames. Table~\ref{tab:tracking} presents the results, showing that leveraging temporal consistency leads to improved tracking performance.

\section{Analysis}
\subsection{Boundary Classifier Analysis} Fig.~\ref{fig:ROC} shows the receiver operating characteristic (ROC) curves for binary boundary classification with a U-Net backbone on the six datasets. The boundary classifier yields outstanding effects on the Fluo-N2DH-SIM+, DIC-C2DH-HeLa, TissueNet, BBBC039, and Fluo-D2DL-HeLa datasets. On PhC-C2DH-U373, the boundary classifier obtains mild accuracy (close to $0.55$) in the AUC-ROC metric. We observe that the irregular cell shapes of this dataset impact the feature quality of the boundary signatures.    

\subsection{Boundary Signature Visualization} 
Fig.~\ref{fig:sigature_visualization} shows some true and false boundary signature examples. We observe that several significant boundary signature patterns can be used to distinguish true and false boundaries. First, we find that true boundary signatures tend to have an X shape (with some arcs). In comparison, false boundary signatures tend to have a T shape or an H shape with some nearly right angles. Second, true boundaries tend to align well between both parts of a boundary signature, while many false boundary signatures have misalignment or a larger distance between the two parts.

\subsection{Boundary Generation Methods Visualization} 
Fig.~\ref{fig:boundary_visualization} shows some visual results by different boundary generation methods. One can see that our method effectively mitigates over-segmentation, under-segmentation, and boundary shape errors in the outputs of the other boundary generation methods.

\begin{table}[t]
\centering
\scriptsize 
\caption[dd]{Tracking performance evaluation using the TRA metric on the video datasets.}
\setlength{\tabcolsep}{1mm}
{

\begin{tabular}{  c |c  |c |c|c }\hline
&  DIC-HeLa
&  Fluo-HeLa 
&  PhC-U373 
& Fluo-SIM+ \\ \hline 

Ceb
& 0.871  & 
0.969 & 0.917 & 0.990  \\
 Ceb+temporal & \textbf{0.889} & \textbf{0.977} & \textbf{0.934} & \textbf{0.991}  \\ \hline

\end{tabular}
\label{tab:tracking}
}
\end{table}

\subsection{Ablation Studies}

\begin{table}[t]
\centering
\scriptsize 
\caption[dd]{Ablation study results. 
}
\setlength{\tabcolsep}{1mm}
{

\begin{tabular}{  c |c c |c c|cc|cc }\hline
&  \multicolumn{2}{c|}{DIC-HeLa}
&  \multicolumn{2}{c|}{Fluo-HeLa} 
&  \multicolumn{2}{c|}{PhC-U373} 
&\multicolumn{2}{c}{Fluo-SIM+} \\ \hline 
  & F1   & AJI & F1  & AJI  
  & F1   & AJI & F1  & AJI \\ \hline
Ceb (WS seed)
& 93.6 & 84.0
 & 95.6 & 92.0 
 & 92.8 & 87.4
 & 97.4 & 84.2  \\
Ceb
& 96.1 & 84.6
 & 96.4 & 92.4  
 &93.2 & 88.7
 & 97.7 & 84.5 \\ \hline
 PM + temporal
 & 96.4 & 84.5
 & 96.4 & 92.5  
 & 93.4 & 89.3
 & 97.4 & 84.2  \\
 Ceb + NMS
 & 94.4 & 83.6
 & 96.4 & 92.4
 & 93.4 & 89.0
 & 97.5 & 84.3  \\
Ceb + temporal
& \textbf{97.1} & \textbf{85.0}
 & \textbf{96.6} & \textbf{92.9}
 & \textbf{93.5} & \textbf{89.5}
 & \textbf{97.8} & \textbf{84.7} \\ \hline

\end{tabular}
\label{tab:ablation_study}
}
\end{table}

\begin{figure}[t]
\centering
\includegraphics[width=0.48\textwidth]{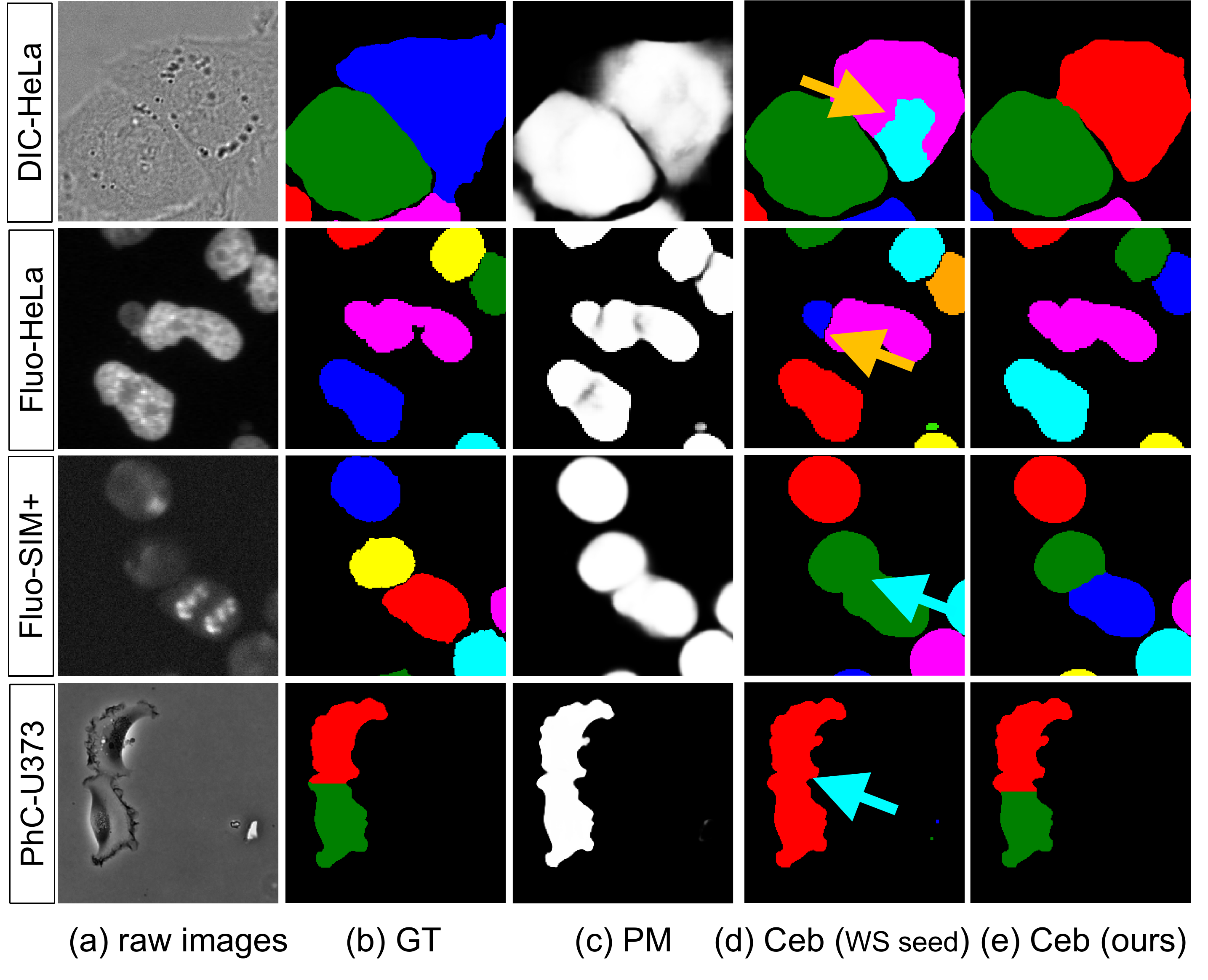}
\caption{Visual comparison between the Ceb version with the original Watershed seed generation (Ceb (WS seed)) and our Ceb method (Ceb (ours)). 
GT denotes ground truth. PM denotes probability map.
Orange arrows indicate over-segmentation errors. Cyan arrows indicate under-segmentation errors.}
\label{fig:experiment_abl}
\end{figure}

To examine the effects of the key components in our approach, we conduct ablation studies using four cell video datasets with a U-Net backbone.

\subsubsection{The Influence of the Seed Generation Method} 
{We replace our seed generation method (see Section~\ref{section:seed_generation}) with the original Watershed seed generation algorithm~\cite{meyer1994topographic}, denoted as \textit{Ceb (WS seed)} in Table~\ref{tab:ablation_study}. As shown in the results, our method \textit{Ceb} consistently outperforms \textit{Ceb (WS seed)}, demonstrating the effectiveness of our proposed seed generation method in capturing all possible cell instances. Fig.~\ref{fig:experiment_abl} shows a visual comparison between our \textit{Ceb} method and \textit{Ceb (WS seed)}.

\subsubsection{The Effect of the Boundary Classifier} 
In the temporal setting, we employ the boundary classifier to determine easy-to-identify instances and create the initial state (see Section~\ref{section:temporal}). We replace our boundary classifier by
the method used in H-EMD~\cite{liang2022h} to create an initial set of selected instances, which selects connected regions from the probability maps (PM) with a threshold of 0.5 such that these regions do not split when the threshold value gets larger, treating such regions as easy-to-identify instances (see \textit{PM + temporal} in Table~\ref{tab:ablation_study}). One can see that our method, \textit{Ceb + temporal}, outperforms \textit{PM + temporal}, demonstrating that our boundary classifier is more effective in determining easy-to-identify cell instances.

\subsubsection{The Effect of the \textbf{SUM} Matching Model} 
We replace our proposed \textbf{SUM} matching model (see Section~\ref{section:temporal}) by the commonly-used Non-Maximum Suppression (NMS)~\cite{he2017mask} selection method. The NMS method selects matching instances in a frame $w$ among the not-yet-selected possible instances in frame $w$ based on their objectness scores using
a greedy strategy (see \textit{Ceb + NMS} in Table~\ref{tab:ablation_study}). The objectness score of a not-yet-selected possible instance $c$ in frame $w$ is defined as the maximum IoU score that $c$ obtains with respect to all selected instances in a considered neighboring frame (say, frame $w+1$) that can possibly form a match with $c$. One can see that our method, \textit{Ceb + temporal}, outperforms \textit{Ceb + NMS}, suggesting that our SUM matching model is more effective in determining optimal matching instance pairs.

{
\section{Limitations}
Our Ceb method is not without limitations. Note that Ceb depends on a full semantic segmentation model (\textit{e.g.}, U-Net) and has five additional components: Seed Generation, Boundary Generation, Boundary Label Assignment (training only), Boundary Signature Extraction, and Boundary Classification. The pipeline of our framework with these five extra components introduces more computational complexity and inherits the limitations associated with probability maps generated by semantic segmentation models. 

\subsection{Computational Complexity}
Table~\ref{tab:computational_complexity} presents a runtime breakdown of the main components in Ceb on the PhC-C2DH-U373 dataset. One can see that Seed Generation and Boundary Signature Extraction take much more time (4.5s and 4.8s) than Semantic Segmentation (1.2s) and Boundary Classification (0.3s). In the current implementation, we use the Python language with loops to process each instance individually in every image, which is time-consuming. However, these computations are independent and can be parallelized for different instances 
concurrently without interference, thus allowing significant speedups if rewritten in C/C++ and integrated with Python via Cython, hence further accelerated on GPUs using CUDA kernels. One example of such implementations can be found in ROIAlign\footnote{https://github.com/multimodallearning/pytorch-mask-rcnn/tree/master/roialign/roi\_align/src} and NMS\footnote{https://github.com/multimodallearning/pytorch-mask-rcnn/tree/master/nms/src} of Mask R-CNN~\cite{he2017mask}, and is beyond the scope of this work.

\subsection{Dependency on Semantic Segmentation}
Our Ceb method relies on a semantic segmentation backbone (e.g., U-Net), and thus depends on the performance of semantic segmentation (i.e., the probability maps). Note that some SOTA methods such as StarDist and CellPose also depend on semantic segmentation for foreground-background pixel distinction, and have the same limitation. Nevertheless, semantic segmentation 
commonly outperforms other methods such as bounding-box based methods (\textit{e.g.}, Mask R-CNN~\cite{he2017mask}) for cell instance segmentation tasks. 
Improving semantic segmentation will improve our method, which focuses on addressing the challenge of distinguishing and clustering background pixels and foreground pixels based on probability maps.

\subsection{Dependency on Watershed}

Our method utilizes a revised Watershed algorithm to generate all potential instance-instance boundaries, and hence depends on the capability of Watershed. 
During instance-wise evaluation, we observed that the potential boundaries thus generated have high recall to cover most ground-truth instance-instance boundaries and, therefore, are effective.

}

\begin{table}[t]
\centering
\scriptsize 
\caption[dd]{Breakdown of computational time of Ceb on a single image of the PhC-C2DH-U373 dataset.}
\setlength{\tabcolsep}{1mm}
{

\begin{tabular}{  c c c}
\toprule
Component & Included in inference & Latency (sec.) \\
\midrule
Semantic Segmentation (U-Net~\cite{ronneberger2015u}) & \checkmark & 1.2 \\
Seed Generation & \checkmark & 4.5 \\
Boundary Generation & \checkmark & 1.1 \\
Boundary Label Assignment &  & 0.6 \\
Boundary Signature Extraction & \checkmark & 4.8 \\
Boundary Classification & \checkmark & 0.3 \\
\bottomrule

\end{tabular}
\label{tab:computational_complexity}
}
\end{table}

\section{Conclusions}
{
Known state-of-the-art cell instance segmentation methods are mainly based on semantic segmentation to distinguish foreground pixels from background pixels. To capture precise cell instances, pixel-wise objectives are commonly used to represent cell instances. However, such pixel-wise representations may overlook geometric properties of cell instances, which may need a structured group of pixels to represent. In this work, we presented a new approach, Ceb, which utilizes cell boundaries in the foreground pixel clustering process. Built on top of existing semantic segmentation backbone models, Ceb transforms the clustering of foreground pixels into a binary boundary classification problem. The boundary classifier is a lightweight CNN based on a novel type of boundary-based features, by sampling pixels from the current foreground-foreground boundary as well as the neighboring background-foreground boundaries. Evaluated on six public cell instance segmentation datasets, Ceb consistently outperforms all the known foreground pixel clustering methods on top of semantic segmentation probability maps. Compared to state-of-the-art cell instance segmentation methods, Ceb obtains comparable or better performances. By incorporating instance temporal consistency in cell videos, our Ceb + temporal method further improves the cell instance segmentation performance. 

\bibliographystyle{ieeetr}
\bibliography{ref}
\end{document}